\documentclass{article}
\usepackage{PRIMEarxiv}

\usepackage{amsmath,amsfonts,bm}

\newcommand{\citep}{\cite}

\def\eqref#1{equation~\ref{#1}}

\def\1{\bm{1}}

\DeclareMathAlphabet{\mathsfit}{\encodingdefault}{\sfdefault}{m}{sl}
\SetMathAlphabet{\mathsfit}{bold}{\encodingdefault}{\sfdefault}{bx}{n}

\usepackage{wrapfig}
\usepackage{hyperref}
\usepackage{url}
\usepackage[utf8]{inputenc} 
\usepackage[T1]{fontenc}    
\usepackage{hyperref}       
\usepackage{url}            
\usepackage{booktabs}       
\usepackage{amsfonts}       
\usepackage{nicefrac}       
\usepackage{microtype}      
\usepackage{lipsum}
\usepackage{fancyhdr}       
\usepackage{graphicx}       
\usepackage{xcolor}
\usepackage{multirow}
\usepackage{adjustbox}
\usepackage{algorithm}
\usepackage{algorithmic}
\usepackage{paralist}
\usepackage{enumitem}
\usepackage{caption}
\usepackage{subcaption}
\newtheorem{theorem}{Theorem}[section]
\newtheorem{proposition}[theorem]{Proposition}

\newtheorem{assumption}[theorem]{Assumption}

\title{WHALE: Towards Generalizable and Scalable World Models for Embodied Decision-making}
\author{%
    Zhilong Zhang\textsuperscript{\rm 1,\rm 2,\rm 3}\thanks{Equal Contribution},~~
    Ruifeng Chen\textsuperscript{\rm 1,\rm 2,\rm 3}\footnotemark[1],~~
    Junyin Ye\textsuperscript{\rm 1,\rm 2,\rm 3}\footnotemark[1],~~
    Yihao Sun\textsuperscript{\rm 1,\rm 2},~~
    Pengyuan Wang\textsuperscript{\rm 1,\rm 2,\rm 3},~~ \\
    \textbf{
    Jingcheng Pang\textsuperscript{\rm 1,\rm 2,\rm 3},~~
    Kaiyuan Li\textsuperscript{\rm 1,\rm 2,\rm 3},~~
    Tianshuo Liu\textsuperscript{\rm 1,\rm 2,\rm 3},~~
    Haoxin Lin\textsuperscript{\rm 1,\rm 2,\rm 3},}~~ \\ 
    \textbf{
    Yang Yu\textsuperscript{\rm 1,\rm 2,\rm 3}\thanks{Corresponding Author},~~ 
    Zhi-Hua Zhou\textsuperscript{\rm 1,\rm 2}} \\
    \textsuperscript{\rm 1}National Key Laboratory for Novel Software Technology, Nanjing University, Nanjing, China\\
    \textsuperscript{\rm 2}School of Artificial Intelligence, Nanjing University, Nanjing, China\\
    \textsuperscript{\rm 3}Polixir Technologies, Nanjing, China\\
     \{zhangzl, chenrf, yejy, sunyh, wangpy, pangjc, liky, liuts, linhx\}@lamda.nju.edu.cn,\\  \{yuy, zhouzh\}@nju.edu.cn
}
\date{}

\begin{document}

\maketitle

\begin{abstract}
World models play a crucial role in decision-making within embodied environments, enabling cost-free explorations that would otherwise be expensive in the real world.
To facilitate effective decision-making, world models must be equipped with strong \textbf{generalizability} to support faithful imagination in out-of-distribution (OOD) regions and provide reliable \textbf{uncertainty estimation} to assess the credibility of the simulated experiences, both of which present significant challenges for prior scalable approaches.
This paper introduces WHALE, a framework for learning generalizable world models, consisting of two key techniques: \emph{behavior-conditioning} and \emph{retracing-rollout}. Behavior-conditioning addresses the policy distribution shift, one of the primary sources of the world model generalization error, while retracing-rollout enables efficient uncertainty estimation without the necessity of model ensembles. These techniques are universal and can be combined with any neural network architecture for world model learning.
Incorporating these two techniques, we present Whale-ST, a scalable spatial-temporal transformer-based world model with enhanced generalizability.
We demonstrate the superiority of Whale-ST in simulation tasks by evaluating both value estimation accuracy and video generation fidelity. Additionally, we examine the effectiveness of our uncertainty estimation technique, which enhances model-based policy optimization in fully offline scenarios.
Furthermore, we propose Whale-X, a 414M parameter world model trained on 970K trajectories from Open X-Embodiment datasets. We show that Whale-X exhibits promising scalability and strong generalizability in real-world manipulation scenarios using minimal demonstrations.
\end{abstract}

\section{Introduction}
\begin{figure*}[t!]
    \centering
    \vspace{-2mm}
    \includegraphics[width=1\linewidth]{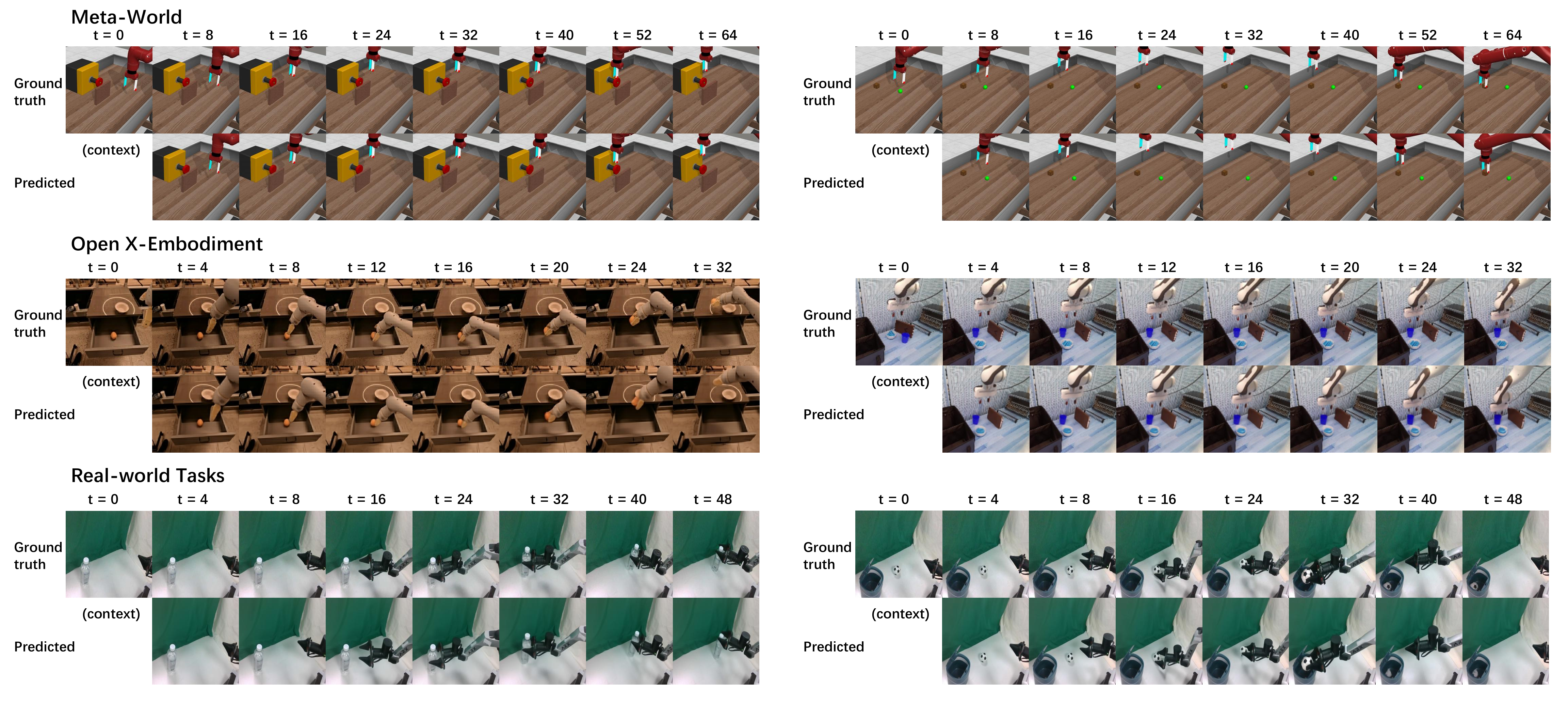}
    \vspace{-3mm}
    \caption{Qualitative evaluation on Meta-World, Open X-Embodiment, and our real-world tasks.}
    \vspace{-3mm}
    \label{fig:qualitative evaluation}
\end{figure*}

Human beings can envision an imagined world in their minds, predicting how different actions might lead to different outcomes~\citep{motion, primary}. Inspired by this aspect of human intelligence, world models~\citep{david2018world} are designed to abstract real-world dynamics and provide such \textit{"what if"} prediction. As a result, embodied agents can interact with world models instead of real-world environments to generate simulation data, which can be used for various downstream tasks, including counterfactual prediction~\citep{adversarial}, off-policy evaluation~\citep{dope}, and offline reinforcement learning~\citep{offline}. 
The requirement to facilitate more effective decision-making presents substantial challenges to the generalizability of world models, an issue that previous approaches have not adequately addressed~\citep{generalist}. Additionally, achieving reliable uncertainty estimation for imagined visual experiences remains a significant challenge, impacting the trustworthy utilization of synthetic data in offline policy optimization~\citep{mopo}. These two unresolved issues impede the further success of world models in supporting decision-making.

In this work, we introduce WHALE (World models with beHavior-conditioning and retrAcing-rollout LEarning), a framework for learning generalizable world models, consisting of two key techniques that can be universally combined with any neural network architecture. First, based on the identification of policy distribution divergence as a primary source of the generalization error, we introduce a \textbf{behavior-conditioning} technique to enhance the generalizability of world models, which builds upon the concept of policy-conditioned model learning~\citep{pcmodel} aiming to enable the model to adapt to different behaviors actively to mitigate the extrapolation error caused by distribution shift.
Furthermore, we propose a simple yet effective technique called \textbf{retracing-rollout}, to enable efficient uncertainty estimation for the model imagination. This approach avoids the necessity of computation-expensive ensembles of the visual world models while providing a reliable uncertainty estimation to facilitate policy optimization in fully offline scenarios. As a plug-and-play solution, the retracing rollout can be efficiently applied to end-effector pose control in various embodiment tasks without necessitating any changes to the training process.

To implement the WHALE framework, we present Whale-ST, a scalable embodied world model based on the spatial-temporal transformer~\citep{st, bruce2024genie}, designed to enable faithful long-horizon imagination for real-world visual control tasks.
To substantiate the effectiveness of Whale-ST, we conduct extensive experiments on both simulated Meta-World~\citep{metaworld} benchmark and a physical robot platform, encompassing a variety of pixel-based manipulation tasks.
Experimental results on the simulated tasks show that Whale-ST outperforms existing world model learning methods in both value estimation accuracy and video generation fidelity.
Moreover, we demonstrate that Whale-ST, based on the retracing-rollout technique, effectively captures model prediction error and enhances offline policy optimization using imagined experiences.
As a further step, we introduce Whale-X, a \textbf{414M-parameter} world model trained on \textbf{970k} real-world demonstrations from Open X-Embodiment datasets~\citep{oxe}. Whale-X serves as a foundational embodied world model for evaluating real-world behaviors. With fine-tuning on a few demonstrations in completely unseen environments and robots, Whale-X demonstrates strong OOD generalizability across visual, motion, and task perspectives. Furthermore, by scaling up the pre-training dataset or model parameters, Whale-X shows impressive scalability during both the pre-training and fine-tuning phases.

The primary contributions of this work are outlined as follows:
\begin{itemize}
\item We introduce WHALE, a framework for learning generalizable world models, consisting of two key techniques: \textit{behavior-conditioning} and \textit{retracing-rollout}, to address two major challenges in the application of world models to decision-making: \textbf{generalization} and \textbf{uncertainty estimation};
\item By integrating these two techniques from WHALE, we propose Whale-ST, a scalable spatial-temporal transformer-based world model designed for more effective decision-making, and further present Whale-X, a 414M-parameter world model pre-trained on 970K robot demonstrations;
\item We conduct extensive experiments to demonstrate the remarkable scalability and generalizability of Whale-ST and Whale-X across both simulated and real-world tasks, highlighting their efficacy in enhancing decision-making.
\end{itemize}

\section{Related Works}
World models have a long research history and recently started to attract significant attention. World models were initially introduced under the name "action models" in simple tasks as components of the decision-making systems~\citep{dyna1990, dyna1991} and were also referred to as "environment models"~\citep{rigter2022rambo}, "dynamics models"~\citep{slbo,mobile}, or simply "models"~\citep{mbpo} in literature. Since the advent of the neural network era, dynamics models have been more widely applied in deep reinforcement learning algorithms to boost learning efficiency, resulting in a series of model-based reinforcement algorithms~\citep{mbpo, pets, mopo, combo, mobile}, while they focused primarily on environment modeling in lower-dimensional proprioceptive state spaces. \cite{david2018world} was the first to propose a general framework for modern world models with high-dimensional visual observations, where a vision module encodes the observed image into a compact latent vector to extract visual information at the current time step and a memory sequence model integrates the historical codes to create a representation that can predict future states. This generic architecture of world models soon achieved a series of notable successes in complex decision-making tasks~\citep{dreamer, mastering, hafner2023mastering, tdmpc, nature}. 

Despite these successes achieved in world model learning, out-of-distribution generalization remains a fundamental challenge for world models, which has yet to be adequately addressed. In contrast to conventional supervised learning settings with no or mild OOD assumption where the target distribution is similar to the training data distribution, world models answer ``what-if" questions: \textit{"What will happen in the environment if the agent makes any possible decisions?"}, which must be highly out-of-distribution. A potential solution to this generalization issue is collecting a larger amount of data to train large world models. Recently, advanced methods have leveraged modern action-conditioned video prediction models~\citep{actionvideo, simple} to model the visual dynamics and pre-train from large-scale video experience data~\citep{bruce2024genie,structure,contextual}. Various sophisticated model architectures have been adopted in these methods, including RNNs~\citep{svg, dreamer, fitvid}, diffusion models~\citep{mcvd, zhu2024sora}, and transformers~\citep{maskvit,ivideogpt, bruce2024genie}. Nevertheless, the available training data for model learning are usually collected by expert or near-expert policies, leading to low data coverage in state-action spaces, which poses challenges to reasoning decision outcomes for suboptimal policies in the learned world models~\citep{generalist}. 

Another line of work investigates the impact of learning methods on the generalizability of world models. For standard maximum likelihood objectives of single-step transitions, the autoregressive rollout errors or value gaps are linked to in-distribution error and policy divergence and are quadratically amplified by the rollout horizon, a phenomenon known as \textit{compounding errors}~\citep{mbpo, errorbound, lambert2022investigating}. To overcome the limitations in the standard MLE learning, a series of improvements have been made, including training multi-step models to reduce rollout errors~\citep{combating, any}, using control objectives to train transition models~\citep{mismatchnomore, simplifymbrl}, adversarially training models for counterfactual target policies~\citep{adversarial}, learning dynamics reward to improve model generalization~\citep{morec}, contrastively learning energy transitions function~\citep{energy}, and incorporating policy information into model inputs to enable test-time model adaptation to target policies~\citep{pcmodel}. Despite the successes in lower-dimensional proprioceptive observation tasks, scaling these methods to large amounts of high-dimensional visual data remains absent. Introducing advanced learning methods suitable for training large world models with large-scale data to improve model generalizability is of unprecedented importance. 

In addition to generalization, another key topic in world models for control is \textit{uncertainty estimation}. It has been shown that offline decision-making within world models learned from partial-coverage data is vulnerable to the exploitation of model prediction errors~\citep{mopo, jinchi}, which requires quantifying the prediction uncertainty of world models and reminding the agent to keep pessimistic about the model uncertainty~\citep{mopo, morel, lurevisiting, mobile}. These typical algorithms estimate the model uncertainty via the ensemble of multiple models learned in parallel~\citep{rafailov2021offline}, which is quite computation-consuming, especially for large-scale tasks and models. Recent works~\citep{kadavath2022language, kuhn2023semantic} adopt the entropy of the categorical token prediction distribution as an uncertainty indicator for large language models (LLMs), though there is limited support for its efficacy in offline model-based control.

\section{Foundations of World Model Learning}
\subsection{Problem Formulation}
A typical formulation of sequential decision tasks is the Markov decision process (MDP)~\citep{mdp,puterman2014markov} specified by the tuple $\mathcal{M}=(\mathcal{S}, \mathcal{A}, r, T^*, \gamma, H, \rho_0)$, where $\mathcal{S}$ is the state space, $\mathcal{A}$ is the action space, $r(s, a)$ is the reward function, $T^*(s^{\prime}|s, a)$ is the real transition probability, $\gamma \in (0,1]$ is the discount factor, $H$ is the decision horizon, and $\rho_0(s)$ is the initial state distribution. In this work, we simply consider the case where $\gamma=1$ and $H<\infty$. 

  In reinforcement learning~\citep{sutton}, the objective is to learn a policy that maximizes the expected return in the MDP, which involves estimating the value of different policies. Specifically, the value of policy $\pi$ is defined as:
\begin{equation}
    V^{\pi}_{T^*} = \mathbb{E}_{\tau_H\sim(\pi,T^*)}\Big[\sum_{t=1}^{H} r(s_t,a_t)\Big],
\end{equation}
where the state-action trajectory $\tau_H=(s_1,a_1,\dots,s_H,a_H)$ and rewards are generated by the rollouts of policy $\pi$ within the dynamics $T^*$. 
A common scenario involves abundant pre-collected experience data, but direct interaction with the environment is either prohibited or costly, necessitating value estimation to be performed offline. 

An environment model $T$ can be explicitly learned from the offline data to imitate the real state transition $T^*$, synthesizing imaginary experiences to simulate real environment interactions. For visual observation tasks, the agent cannot directly observe the states; instead, it receives high-dimensional images within the observation space $\mathcal{O}$, which normally introduces redundant information and partial observability. The visual world models~\citep{david2018world} usually learn a vision module $E_\theta: \mathcal O\rightarrow \mathcal S$ to extract a compressed representation $z_t=E_\theta(o_t)$ from the current frame observation $o_t$, and use a sequence model to integrate the latent representations $z_{1:t}$ from past frames as well as actions $a_{1:t}$ to overcome the partial observability for future prediction. This architecture of world models empowers autoregressive imagination for any given policy, which enables policy evaluation and improvement without real-world interactions.

Assume that $ V^\pi_{T}$ is the value estimated within the model $T$, the environment model error induces a value gap $|V^\pi_{T^*}-V^\pi_{T}|$ for the policy $\pi$. 
If the model is globally accurate, the value gap will diminish for any policy. However, offline experiences are often collected by a narrow range of policies (e.g., near-expert policies). Therefore, the learned environment models are likely unfamiliar with the outcomes of novel decision patterns and are expected to generalize beyond the training experiences for counterfactual reasoning to evaluate diverse policies. 

\subsection{Generalizability of World Models}

The common learning methods for world models regard the transition learning as a standard supervised learning problem, minimizing the negative log-likelihood (NLL) of the single-step transition probabilities over the pre-collected trajectories in a teacher-forcing manner, i.e., 
$$
\min_{T}\mathbb E_{\mu\sim\Pi}\mathbb E_{\tau_H\sim(\textcolor{red}{\mu,T^*})}\frac{1}{H}\sum_{h=1}^H \mathbb -\log T(o_h|\tau_{h-1}) \ \Longleftrightarrow\ \min_{T}l_{\mathrm{KL}}(T;\Pi),
$$

where (sub-)trajectory $\tau_h=(o_1,a_1,o_2,\dots, o_h,a_h), 1\leq h \leq H$ is generated by interaction of a behavior policy $\mu$ with the real dynamics $T^*$, and behavior $\mu$ is assumed to be sampled from a behavior policy distribution $\Pi$. Minimizing the NLL equals minimizing the KL divergence loss
$\displaystyle l_{\mathrm{KL}}(T;\Pi)=\mathbb E_{\mu\sim\Pi}\mathbb E_{\tau_H\sim(\mu, T^*)}\frac{1}{H}\sum_{h=1}^HD_{\mathrm{KL}}(T^*(\cdot|\tau_{h-1}),T(\cdot|\tau_{h-1})).$
The learned world models are usually utilized to evaluate any target policy $\pi$ by simulating trajectories in an autoregressive manner:
$$
V^\pi_T = \mathbb E_{\tau_H\sim (\textcolor{red}{\pi, T})} \Big[\sum_{t=1}^Hr(o_t,a_t)\Big],
$$
where the trajectory simulation distribution deviates from the training distribution.

In classical sequential modeling tasks like sentence generation and translation, the distribution shift from teacher-forcing training to autoregressive generation diminishes as the model accuracy improves, which therefore does not lead to significant negative impacts. For world model learning, however, the distribution shift results from both the model prediction inaccuracy and the divergence between the target policy and behavior policies, exacerbating the evaluation inaccuracy:
\begin{equation}
    \Big|V^\pi_T - V^\pi_{T^*}\Big| \leq 2R_{\max}\underbrace{H^2}_{\textcolor{blue}{\mathrm{teacher-forcing}}}\Big(\underbrace{\sqrt{2~l_\mathrm{KL}(T;\Pi)}}_{\mathrm{in~distribution~error}}+\underbrace{L\cdot W_1(d^{\pi},d^{\Pi})}_{\textcolor{red}{\mathrm{policy~divergence}}}\Big),
    \label{eq:error_dec}
\end{equation}
where a distribution shift term induced by the policy divergence~\footnote{Here $W_1(d^\pi,d^\Pi)$ is the Wasserstein-1 distance between the $\pi$-induced trajectory distribution $d^\pi(\tau)$ and the behavior trajectory distribution $d^\Pi(\tau)=\mathbb E_{\mu\sim\Pi}[d^\mu(\tau)]$, and $L$ is the Lipschitz constant of model loss w.r.t. the trajectory, adapted from \cite{pcmodel}.} occurs in addition to the KL training loss, further amplified by an $H^2$ factor caused by the supervised teacher-forcing learning. 
Even if the world model perfectly models the training distribution, i.e. $l_{\mathrm{KL}}(T;\Pi)=0$, the variation of the target policies could also significantly shift the trajectory simulation distribution to those large error areas, resulting in degenerative generalizability. 

The generalization issue of world model learning has been noticed as a severe challenge even when large expert training data is available \cite{generalist}. Previously, solutions have been proposed. 
\begin{itemize}
\item Replacing the teacher-forcing objective by the distribution matching solves the compounding error \cite{errorbound}, which reduces the term $H^2$ to $H$. 
\item Energy-based model can help reduce the in distribution error \cite{energy}, which waives the need of fitting ill-shaped transition functions by neural networks.
\item The adversarial counterfactor learning \cite{adversarial} alleviate the policy divergence issue. 
\item Learning an environment reward to constrain the world model can surprisingly help generalizes out of the data distribution \cite{morec}.
\end{itemize}
Meanwhile, the above methods are all based on adversarial learning, which is hard to scaling up currently. In \cite{pcmodel}, a new way to enhance the generalizability was found, which learns to generalize during training by enforcing a policy condition. This paper inherits and improves this method.

\section{Learning Generalizable World Models for Embodied Decision-making}

Sequential decision-making within world models often necessitates agents to explore out-of-distribution (OOD) regions that go beyond the training dataset. This requires the world models to exhibit strong generalization capabilities, enabling it to make accurate predictions that closely mirror real-world dynamics. Meanwhile, reliably quantifying the prediction uncertainty, especially for transitions outside the world models' effective generalization range, is essential for robust decision-making, which prevents offline policy optimization from exploiting erroneous model predictions. 
In consideration of these issues, we propose \textbf{WHALE}, a framework for learning generalizable world models with enhanced generalizability and efficient uncertainty estimation. 

\subsection{Behavior-conditioning for Generalization} 
 According to the error decomposition in Eq (\ref{eq:error_dec}), the generalization error of the world model primarily arises from error compounding caused by policy divergence. To reduce this error, one possible solution to this policy generalization issue is to embed the behavior information into the world model, allowing the model to actively recognize the behavior patterns of the policies and adapt to the policy-induced distribution shift~\citep{pcmodel}. This adaptation effect has been shown to reduce model generalization error caused by policy divergence, i.e. the last term in Eq (\ref{eq:error_dec}). For further analysis, please refer to Appendix \ref{app:analysis}. Building upon the concept of behavior-conditioning, we introduce a learning objective to obtain behavior embeddings from training trajectories and integrate the learned embeddings. 

We would like to extract the decision patterns within training trajectories $\tau_H$ into a behavior embedding, reminiscent of the maximization of the evidence lower bound (ELBO) of the trajectory likelihood conditioned on the history $\tau_h$~\citep{reason, flow2control, opal}:
\begin{equation}
    \log P(\tau_H|\tau_h)\geq \mathbb E_{q_\phi(z|\tau_H)}\sum_{t=h}^H\log\pi_w(a_t|o_t,\tau_{t-1},z) - D_{\mathrm{KL}}(q_\phi(z|\tau_H)||p_\psi(z|\tau_h)) + Const,
\end{equation}
where $q_\phi(z|\tau_H)$ denotes the posterior encoder, encoding the whole trajectory $\tau_H$ into a latent variable $z$; $\pi_w(a_h|o_h,\tau_{h-1},z)$ denotes the decoder, which recovers the decision action from the latent variable $z$ and the up-to-date history $(\tau_{h-1}, o_h)$; $p_\psi(z|\tau_h)$ denotes the prior predictor, which allows the prediction of $z$ based on the history $\tau_h$. The information bottleneck requires the learned variable $z$ to effectively capture the decision pattern within the trajectory, embedding the information about the corresponding behavior policy. Following this argument, we propose to learn the behavior embedding by maximizing the ELBOs over $H$ decision steps and adjusting the amount of KL constraints similar to $\beta$-VAE~\citep{betavae}:
\begin{equation}
    \mathcal L(w,\phi,\psi) = \mathbb E_{\tau_H\sim\mathcal D}\Big[ \mathbb E_{q_\phi(z|\tau_H)}\big[ -\sum_{h=1}^H\log\pi_w(a_h|o_h,\tau_{h-1},z) \big] + \beta \sum_{h=1}^H D_{\mathrm{KL}}(q_\phi(z|\tau_H)|| p_\psi(z|\tau_h)) \Big],
    \label{eq:emb_learning}
\end{equation}
here the KL terms constrain the embedding predictions from sub-trajectories up to each time step $h$, encouraging them to approximate the posterior encoding. This ensures that the representation remains policy-consistent, meaning that trajectories generated by the same policy exhibit similar behavioral patterns and, consequently, similar representations.

The learned prior predictor $p_\psi$ is then used to obtain behavior embeddings $z_h$ from history $\tau_h$ for behavior-conditioning during world model learning, where the behavior embeddings are accepted as additional covariates for future prediction:
$$
\min_{T}\mathbb E_{\mu\sim\Pi}\mathbb E_{\tau_H\sim(\mu,T^*)}\frac{1}{H}\sum_{h=1}^H \mathbb E_{\textcolor{red}{z_h\sim p_\psi(\cdot|\tau_h)}} -\log T(o_{h+1}|\tau_{h},\textcolor{red}{z_{h}}).
$$
When rolling out target policies or executing action sequences within the learned world models, the prior predictor infers the latent behavior intentions from interaction history, enabling the autoregressive generation process to adjust to the target distribution on the fly for future imagination adaptively.

\subsection{Retracing-rollout for Uncertainty Estimation}
\begin{figure*}[!t]
    \centering
    \includegraphics[width=0.95\linewidth]{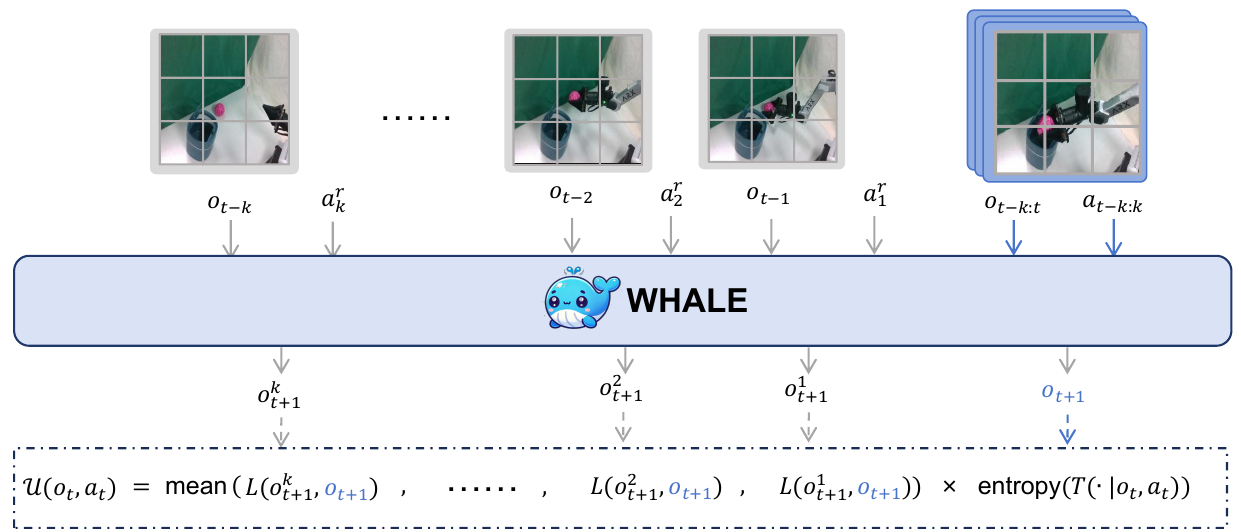}
    \caption{Illustration of retracing-rollout uncertainty qunatifier. }
    \vspace{-3mm}
    \label{fig:uncertainty}
\end{figure*}
Agents can take any actions in world models to generate an imagined future, which may significantly diverge from the offline dataset. As a result, world models inevitably produce inaccurate and unreliable samples. Previous works have demonstrated, both theoretically and experimentally, that if we use model-generated data without restrictions, the performance of the policy can be severely undermined~\citep{mopo,jinchi}. Therefore, uncertainty estimation is essential for world models, as it can indicate when to trust our models. 

Previous ensemble-based uncertainty estimation methods for world models often require training multiple models~\citep{rafailov2021offline,mobile}, making them computationally expensive, especially for large-scale tasks and complex models. Beyond these, any-step uncertainty estimation ~\citep{lin2024anystepdynamicsmodelimproves} has emerged as a computationally efficient alternative, leveraging discrepancies in predictions across historical information of varying lengths, without relying on ensemble. However, any-step is specifically designed for recurrent neural networks and cannot be directly applied to transformers, limiting its scalability and applicability in real-world tasks. To overcome this limitation, we introduce a novel uncertainty estimation method, \textbf{retracing-rollout}, inspired by the concept of variable-length history in any-step. The core innovation of retracing-rollout lies in the introduction of \textbf{retracing-action}, which leverages the semantic structure of the action space in embodied control, enabling more accurate and efficient uncertainty estimation for transformer-based world models. 

We begin by introducing retracing-action. Concretely, retracing-action serves as an equivalent substitute for any given action sequence. For an action sequence $a_{1:k}$, its corresponding retracing-action is defined as $a^r_k$, with the goal that the outcome $o_{k+1}$ produced by the robotics executing action sequence $a_{1:k}$ from any observation $o_1$ nearly identical to the outcome $o^r_{k+1}$ produced when executing the retracing-action $a^r_k$. Benefiting from the semantic structure of the action space in embodied control, the retracing-action is computationally feasible for end-effector pose control. For instance, in the Open X-embodied dataset, the action space is defined by a 7-dimensional vector that controls the end-effector. The first three dimensions represent the changes in the gripper position ($\Delta x$, $\Delta y$, $\Delta z$), the next three represent the changes in wrist orientation ($\Delta$roll, $\Delta$pitch, $\Delta$yaw), and the final dimension determines whether the gripper opens or closes. Therefore, the retracing-action can be directly computed using Eq (\ref{eq:retrace_action}), where $a_i^{(j)}$ represents the value of the $j$-th dimension of the action $a_i$. 
\begin{equation}
\label{eq:retrace_action}
     a^r_k= \left(\sum_{i=1}^k a_i^{(0)},\cdots,\sum_{i=1}^k a_i^{(5)}, a_k^{(6)}\right).
\end{equation}

The next concept is retracing-rollout. Given a retracing step $k$, the process begins by backtracking to $o_{t-k}$ as the starting frame for a rollout. The correspounding retracing-action $a_k^r$ for action sequence $a_{t-k:k}$ is executed from $o_{t-k}$, yielding the corresponding outcome $o_{k+1}$. In practice, to prevent multi-step cumulative retracing-action $a_k^r$ from exceeding the action-space range, $a_k^r$ is divided into $k$ steps. In each step, the first six dimensions are set to $\frac{1}{k}a_k^{r(i)}$, while the last dimension, $a_k^{r(6)}$, remains constant, allowing us to achieve desired outcomes through a multi-step rollout. 

Finally, we present the retracing-rollout uncertainty quantifier, as illustrated in Figure~\ref{fig:uncertainty}. To estimate uncertainty for $(o_{t}, a_{t})$, various retracing steps are set to generate corresponding retracing-rollout predictions. We then calculate the feature-level disagreement, represented as perceptual loss, between outputs generated from retracing-rollout and the output without retracing-rollout. Additionally, we incorporate the predictive entropy of the dynamics model $\text{entropy}(T(\cdot|o_t, a_t))$, which is typically used to measure the confidence of the model itself. By multiplying the average disagreement and predictive entropy, we derive the final uncertainty estimation result. Notably, retracing-rollout does not require any modifications during the training phase, which significantly reduces the computational overhead compared to ensemble-based methods. 

\section{Practical Implementation of WHALE}
In this section, we present an instantiation of WHALE, Whale-ST, a scalable world model with ST-transformer architecture. As a further step, we introduce Whale-X, a 414M-parameter world model trained on 970k real-world demonstrations from Open X-Embodiment datasets.
\subsection{Instantiation with ST-transformer architecture: Whale-ST}
\begin{figure*}[!t]
    \centering
    \includegraphics[width=0.95\linewidth]{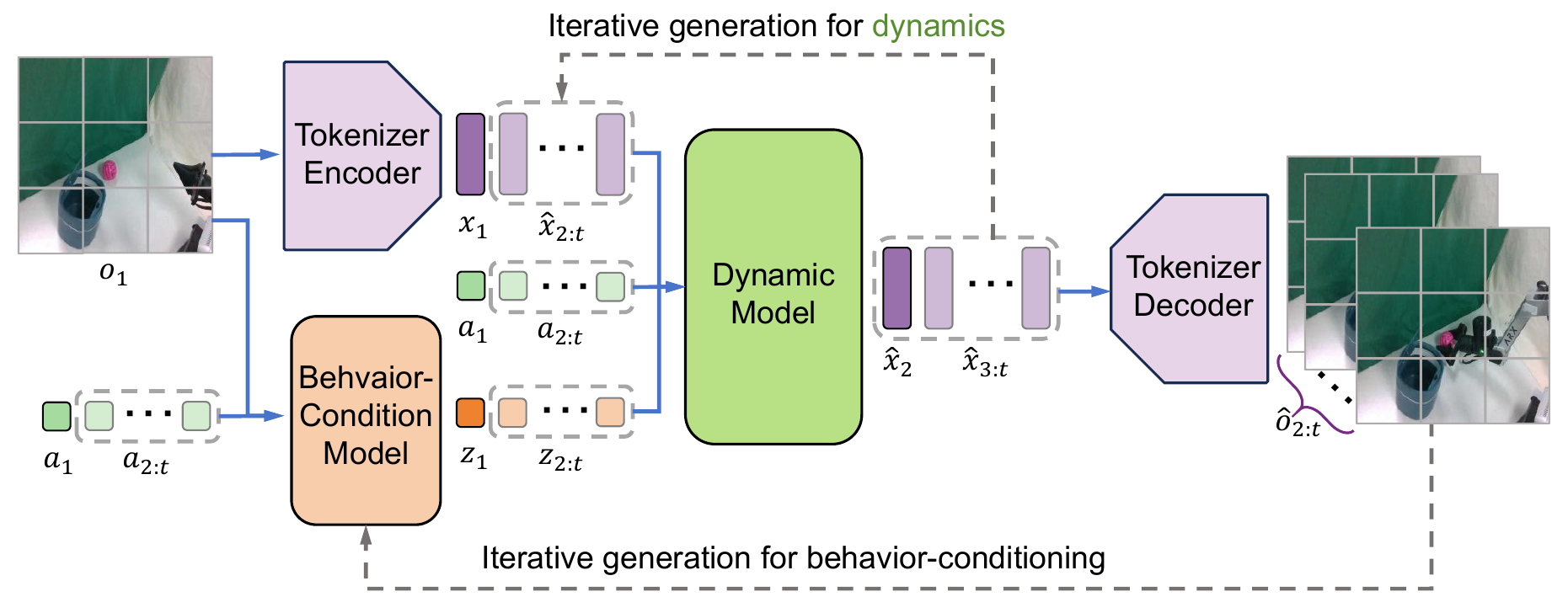}
    \caption{Overall architecture of Whale-ST. The behavior-conditioning model encodes the observation and action subsequences into behavior embedding $z_i$, which are then passed to the dynamics model along with observation tokens and actions to generate the next token predictions $\hat x_{i+1}$. The predicted observation tokens are subsequently fed into the dynamics model for further predictions autoregressively and decoded into observation predictions to obtain later behavior embeddings.}
    \vspace{-3mm}
    \label{fig:architecture}
\end{figure*}

Figure \ref{fig:architecture} illustrates the overall architecture of Whale-ST. Specifically, Whale-ST comprises three main components: behavior-conditioning model, video tokenizer, and dynamics model. Inspired by previous works~\citep{bruce2024genie}, these modules utilize a spatial-temporal transformer (ST-transformer) architecture. Within this framework, each token is designed to attend only to other tokens in the current frame and those at corresponding positions in prior frames. Additionally, Whale-ST can generate all tokens for the next frame in parallel at one time. These designs significantly simplify the computational demands from a quadratic to a linear dependency relative to sequence length, reducing both the memory usage and computational costs of the model training while increasing model inference speed.

\subsubsection{Behavior-conditioning Model Learning}
Behavior-conditioning model comprises a CNN-based visual encoder $v_\theta$, ST-Transformer-based posterior model $q_\phi$, prior model $p_\psi$, and reconstruction model $\pi_\omega$. Given an input image sequence, $v_\theta$ first converts it into tokens by patchifying the images. These tokens are then processed by $q_\phi$, $p_\psi$, and $\pi_\omega$, which produce the posterior representations $z_H$, prior representations $z_h$, and reconstructed actions $a_h$, respectively. As a results, we can train the behavior-conditioning model using the loss function defined in Eq (\ref{eq:emb_learning}). For behavior embeddings, we employ two-hot encoding due to its strong expressive capacity and stable training process, as noted in \cite{dreamer}.

\subsubsection{World Model Learning}
World models typically consist of an observation encoder that encodes the raw observation into a compact representation and a dynamics model that predicts future transitions within this representation space~\citep{david2018world}. In this work, we adopt a tokenizer based on VQ-VAE~\citep{van2017neural} as the encoder to discretize observations into tokens and train a dynamics model at the token level.

Specifically, the video tokenizer $e_\theta$ is composed of an encoder $E_\theta$ and a decoder $D_\theta$, where the encoder $E_\theta$ compresses video input into a sequence of tokens, while the decoder $D_\theta$ is capable of reconstructing the original video from these tokens. 
This tokenizer is trained with the standard VQ-VAE loss $\mathcal{L}_\text{tok}(\theta)$ \label{eq:tok_learning}, which is a combination of a $L_1$ reconstruction loss, a codebook loss, and a commitment loss. 

After training the tokenizer, we embed the policy information into the dynamics model learning process. The key distinction from standard dynamics model learning is that Whale-ST additionally incorporates a behavior-conditioning $z_h$ inferred by the prior predictor $p_\psi$. In this phase, for each input trajectory segment $\tau_H$, the video tokenizer first converts it into a sequence of tokens $x_H=((x_1^{(1)},\cdots,x_1^{(N)}),(x_2^{(1)},\cdots,x_2^{(N)}),\cdots,(x_H^{(1)},\cdots,x_H^{(N)}))$, where $x_i^{(j)}$ represents the $j$-th token of the $i$-th frame. Consequently, the training objective of the dynamics model is to maximize the log-likelihood of the tokens $x_{h+1}$ for the next frame $s_{h+1}$, conditioned on the history tokens $x_{0:h}$, history actions $a_{0:h}$ and the behavior-conditioning $z_h = p_\psi(\tau_h)$:
\begin{equation}
    \mathcal L_\text{dyn}(\theta) = \mathbb E_{\tau_H\sim\mathcal D}\big[ -\sum_{h=1}^H\log P_\theta(x_{h+1}|x_{1:h}, a_{1:h}, z_h) \big],
    \label{eq:dyn_learning}
\end{equation}

Intuitively speaking, Whale-ST not only accepts history as a direct feature to predict transitions but also infers the latent decision intention from the history to enable test-time adaptation to the induced distribution shift.

\subsection{Pre-training on real-world robot demonstrations: Whale-X}
We introduce Whale-X, a 414M-parameter world model pre-trained on 970K real-world robot demonstrations from the Open X-Embodiment dataset. The entire dataset is utilized to pre-train both the behavior-conditioning model and the video tokenizer, while a focused subset of the data is used to pre-train the dynamics model. Whale-X serves as a foundational embodied world model for assessing real-world behaviors, capable of generating realistic, controllable video trajectories that align closely with specified actions, as illustrated in Figure~\ref{fig:qualitative evaluation}.

\textbf{Data.}  Whale-X is pre-trained on the Open X-Embodiment dataset~\citep{oxe} (OpenX), a comprehensive dataset consists of more than 70 individual robot datasets, with more than 2M robot trajectories pooled into a coherent and easy-to-use data format in a large community effort. Follow~\cite{openvla}. our pre-training dataset collection includes 27 datasets, with a total scale of 970k demonstrations, we list our used data mixture and weights in Table~\ref{tab:data_mix}. To train a world model focused on tabletop tasks, we extract data related to tabletop tasks from the dataset that features similar camera positions (the bolded tasks in Table \ref{tab:data_mix}) to train the dynamics model, while the video tokenizer and behavior-conditioning model are trained on the full OpenX dataset.

\section{Experiment}

We conduct extensive experiments on both simulated tasks and real-world tasks. The experimental design is primarily designed to answer the following key questions:
\begin{itemize}[leftmargin=1em]
    \item How does Whale-ST perform compared with other baselines on simulated tasks? Are behavior-conditioning and retracing-rollout techniques effective?  (Section~\ref{sec:exp_sim})
    \item How does Whale-X perform on real-world tasks? Can Whale-X benefit from pre-training on internet-scale data? (Section~\ref{sec:exp_real})
    \item How is the scalability of Whale-X? Does increasing the model capacity or pre-training data improve performance on real-world tasks? (Section~\ref{sec:exp_scale})
\end{itemize}

\subsection{Whale-ST on Simulated Tasks}
\label{sec:exp_sim}
\subsubsection{Experiment Setups}
 We conduct our simulated task experiments on the Meta-World~\citep{metaworld} benchmark, which offers a diverse set of vision-based manipulation tasks. In this experiment, we construct a training dataset with 60k trajectories collected from 20 tasks. The model learning algorithms are required to use all the data for training from scratch. Afterward, we conduct model evaluation and uncertainty estimation experiments to validate the effectiveness of behavior-conditioning and retracing-rollout, respectively. More detailed information about data collection can be found in Appendix~\ref{data:sim}. 

\subsubsection{Model Evaluation}
\textbf{Baselines.} We compare Whale-ST against several world model learning baselines, including \begin{inparaenum}[(1)]
  \item \textbf{FitVid}~\citep{fitvid}, a variational-based world model that can fit large diverse video datasets.
  \item \textbf{MCVD}~\citep{mcvd}, a diffusion-based world model that can perform video generation conditioning on different subsets of video frames and actions.
  \item \textbf{DreamerV3}~\citep{hafner2023mastering}, a recurrent world model that outperforms specialized methods across diverse control tasks. 
  \item \textbf{iVideoGPT}~\citep{ivideogpt}, a scalable transformer-based world model that achieved state-of-the-art results in video generation and embodied control tasks.
\end{inparaenum} Complete descriptions and implementation details are provided in Appendix~\ref{impl:baselines}.

\textbf{Evaluation Metrics.} We assess the performance of world models from two perspectives: \begin{inparaenum}[(1)]
  \item\textbf{Value estimation accuracy}. Verifies whether the model can correctly estimate the value of a given action sequence, in terms of Value Gap, Return Correlation, and Regret~\citep{dope}. 
  \item\textbf{Video fidelity}. Measures the quality of video trajectory generation, in terms of FVD~\citep{unterthiner2018towards}, PSNR~\citep{huynh2008scope}, LPIPS~\citep{zhang2018unreasonable}, and SSIM~\citep{wang2004image}. 
\end{inparaenum} More detailed information about evaluation metrics is provided in Appendix~\ref{impl:metrics}.

\textbf{Task Results.} Table \ref{tab:value_comparison} presents the results for value prediction accuracy, where Whale-ST demonstrates exceptional performance across all three metrics. At a $64 \times 64$ resolution, Whale-ST achieves a value gap closely aligning with DreamerV3's top score. Moreover, it surpasses all other models on return correlation and regret@5, highlighting minimal prediction errors. When tested at a higher resolution of $256 \times 256$, the performance of Whale-ST further improves, achieving the smallest value gap and the highest return correlation, reflecting an even finer understanding of environment dynamics. With the lowest regret, Whale-ST provides high-confidence guidance for decision-making. Furthermore, Table \ref{tab:meta_comparison} presents the results for video fidelity, showing that Whale-ST consistently outperforms all other methods across metrics related to video fidelity, with a notable advantage in FVD. 

\textbf{Ablation Study.} To validate the effectiveness of behavior-conditioning, we conduct ablation experiments at a resolution of $256\times256$, as shown in the last two rows of Table~\ref{tab:value_comparison} and \ref{tab:meta_comparison}. Incorporating behavior-conditioning markedly improves value prediction accuracy in world models, reducing the value gap by 18\% and regret by 22\%. Furthermore, behavior-conditioning consistently enhances video fidelity across all metrics.

\begin{table}[h!]
\centering
\begin{tabular}{lccccc}
\toprule
\textbf{Meta-World} & \#Params & Value Gap$\downarrow$ & Return Correlation$\uparrow$ & Regret@5$\downarrow$ \\
\midrule
\multicolumn{5}{c}{\textit{64$\times$64 resolution}} \\
\midrule
FitVid & 143M & 18.2 & 0.64 & 22.0 \\
MCVD & 53M & 20.6 & 0.72 & 12.2 \\
DreamerV3 & 44M & \textbf{10.0} & 0.70 & 16.5 \\
iVideoGPT & 63M & 15.9 & 0.62 & \textbf{7.2} \\
Whale-ST (ours) & 51M & \textbf{10.3 $\pm$ 0.8} & \textbf{0.77$\pm$0.01} & \textbf{7.3$\pm$1.2} \\
\midrule
\multicolumn{5}{c}{\textit{256$\times$256 resolution}} \\
\midrule
DreamerV3 & 61M & 11.7 & 0.72 & 11.5 \\
Whale-ST (w/o bc) & 63M & 11.8$\pm$0.4 & 0.84$\pm$0.01 & 5.0$\pm$0.3 \\
Whale-ST (ours) & 63M & \textbf{9.7$\pm$0.1} & \textbf{0.86$\pm$0.01} & \textbf{3.9$\pm$0.3} \\
\bottomrule
\end{tabular}
\caption{Value prediction accuracy comparison on Meta-World benchmark with various models.}
\vspace{-3mm}
\label{tab:value_comparison}
\end{table}

\begin{table}[h!]
\centering
\begin{tabular}{lccccc}
\toprule
\textbf{Meta-World} & \#Params & FVD$\downarrow$ &  PSNR$\uparrow$ & SSIM$\uparrow$ & LPIPS$\downarrow$  \\
\midrule
\multicolumn{6}{c}{\textit{64$\times$64 resolution}} \\
\midrule
FitVid & 143M& 154.6 & 23.7 & 90.3 & 6.5  \\
MCVD & 53M & 272.8& \textbf{29.7}& 92.3& 4.0 \\
DreamerV3 & 44M & 142.7 & 27.6 & 92.1 & 4.3  \\
iVideoGPT & 63M &  115.7& 28.5&92.8  &  4.5 \\
Whale-ST (ours)& 51M  & \textbf{38.5$\pm$2.6} & 28.8$\pm$0.0 & \textbf{93.5$\pm$0.1} &\textbf{3.7$\pm$0.1} \\
\midrule
\multicolumn{6}{c}{\textit{256$\times$256 resolution}} \\
\midrule
DreamerV3 & 61M & 112.4&  26.2&  91.7& 8.5 \\
Whale-ST (w/o bc) & 63M & 32.0$\pm$0.4 & 28.9$\pm$0.2  & 94.6$\pm$0.1 &4.6$\pm$0.1 \\
Whale-ST (ours) & 63M& \textbf{28.2$\pm$3.6} & \textbf{29.2$\pm$0.2} & \textbf{95.0$\pm$0.1} &\textbf{4.3$\pm$0.1}\\
\bottomrule
\end{tabular}
\caption{Video fidelity comparison on Meta-World benchmark with various models.}
\vspace{-3mm}
\label{tab:meta_comparison}
\end{table}

\subsubsection{Uncertainty Estimation}

\textbf{Baselines.} We compare the retracing-rollout technique with two baseline methods for uncertainty estimation: \begin{inparaenum}[(1)] \item \textbf{Entropy-based Method}: We use the predictive entropy of the transformer-based dynamics model to quantify uncertainty, following approaches similar to those in~\cite{kadavath2022language, kuhn2023semantic}. \item \textbf{Ensemble-based Method}: We independently train three dynamics models and estimate uncertainty by measuring the pixel-level disagreement between images generated by each model, following methods described in~\cite{yu2020mopo}. Detailed baseline implementations are provided in Appendix~\ref{impl:baselines}.
\end{inparaenum}

\textbf{Evaluation Metrics.} We evaluate uncertainty estimation methods from two perspectives: \begin{inparaenum}[(1)]
\item\textbf{Model Error Prediction}. We evaluate uncertainty by
treating uncertainty estimation as the problem of predicting whether to trust the generated trajectory with given observations and actions.  Therefore, we compute the correlation between the estimated uncertainty and the perceptual model error, defined as the perceptual loss between real observations and generated samples. A higher correlation suggests a more reliable uncertainty estimation. \item\textbf{Offline Reinforcement Learning}. According to~\cite{mopo}, offline model-based reinforcement learning (offline MBRL) relies on estimating the fidelity of generated samples to support conservative policy learning, where more accurate uncertainty estimates lead to improved and more stable policies. Consequently, we also evaluate how effectively the estimated uncertainty contributes to model-based policy optimization in offline MBRL. Following previous works~\citep{mopo, rigter2022rambo}, we use the estimated uncertainty to reshape the reward, i.e., $r^\prime(s,a)=r(s,a) - \alpha \mathcal{U}(s,a)$, where $r(s,a)$ is the original reward and $\alpha$ is a hyperparameter determines the extent of the uncertainty penalty. \end{inparaenum}
Appendix~\ref{impl:metrics} provides more detailed information about evaluation metrics.

\begin{figure*}[htbp]
    \centering
    \vspace{-2mm}
    \includegraphics[width=0.95\linewidth]{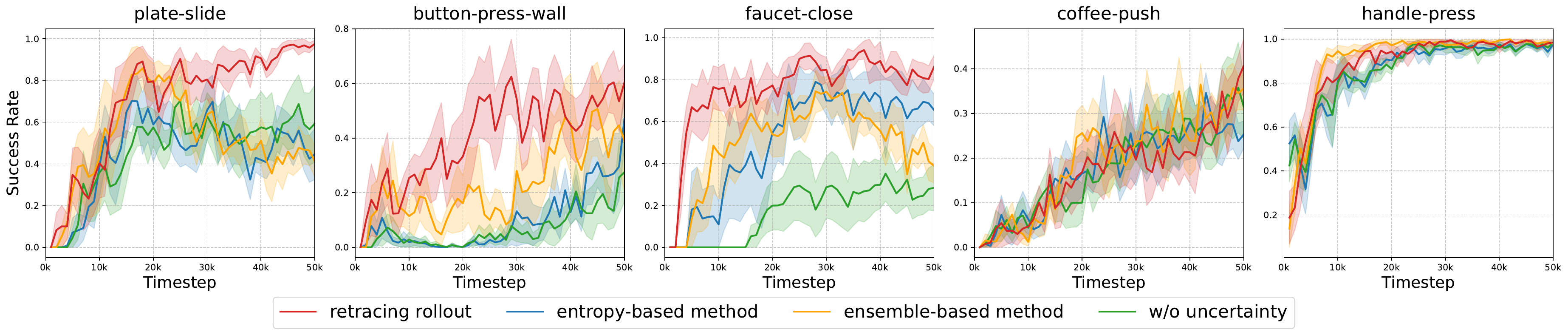}
    \caption{Offline reinforcement learning with different uncertainty estimation methods.}
    \vspace{-3mm}
    \label{fig:offline_rl}
\end{figure*}

\textbf{Task Results.} Table \ref{tab:uncertainty} demonstrates the model error prediction results, showing that retracing rollout consistently outperforms other baselines in all 5 tasks. Specifically, compared to the ensemble-based method, retracing rollout achieves a significant 500\% improvement, and compared to the entropy-based method, it shows a 50\% increase. Additional results are available in Appendix \ref{app:uncertainty_quanti}. Furthermore, Figure \ref{fig:offline_rl} displays the offline MBRL outcomes, where retracing rollout achieves superior convergence performance and training stability in three of the five tasks while performing comparably in the remaining two. Notably, in \textit{faucet-close} and \textit{plate-slide} tasks, retracing rollout is the only method capable of stable convergence, whereas other methods experience varying degrees of performance decline in later training stages. Additionally, compared to learning without an uncertainty penalty, all methods deliver significant improvements, further confirming the importance of uncertainty estimation in policy learning.
\begin{table}[htbp]
    \centering
    \begin{tabular}{lccc}
    \toprule
    \textbf{Tasks} & Ensemble-based & Entropy-based & Retracing Rollout (ours) \\
    \midrule
    plate-slide & 0.14$\pm$0.01 & 0.31$\pm$0.08 & \textbf{0.55$\pm$0.09} \\
    button-press-wall & 0.11$\pm$0.01 & 0.31$\pm$0.03 & \textbf{0.52$\pm$0.07} \\
    faucet-close & 0.06$\pm$0.05 & 0.40$\pm$0.03 & \textbf{0.70$\pm$0.01} \\
    coffee-push & 0.06$\pm$0.01 & \textbf{0.49$\pm$0.06} & \textbf{0.53$\pm$0.04} \\
    handle-press & 0.10$\pm$0.07 & 0.25$\pm$0.06 & \textbf{0.45$\pm$0.08} \\
    \midrule
    \textbf{Average} & 0.10$\pm$0.03 & 0.35$\pm$0.05 & \textbf{0.55$\pm$0.06} \\
    \bottomrule
    \end{tabular}
    \caption{Correlation between estimated uncertainty and perceptual model error.}
    \vspace{-3mm}
    \label{tab:uncertainty}
\end{table}

\subsection{Whale-X on Real-world Tasks}
\label{sec:exp_real}

\subsubsection{Experiment Setups} 
To evaluate the generalizability of Whale-X in the physical world, we conduct comprehensive real-world experiments on ARX5 robotic platform. The evaluation tasks differ significantly from the pre-training data, in terms of the robotic platform, camera angles, and background visual information, posing considerable challenges for world models.

We carefully collect a limited dataset for fine-tuning, consisting of 60 trajectories for each of the four tasks:  \textit{open bin}, \textit{push plate}, \textit{throw ball}, and \textit{move bottle}.  Following this, we designed several challenging unseen tasks for testing, with a focus on evaluating the model from the perspectives of \textit{visual generalization}, \textit{motion generalization}, and \textit{task generalization} perspectives. Further details on the fine-tuning and data collection process can be found in Appendix~\ref{app:finetune} and Appendix~\ref{app:data collection} respectively.

\subsubsection{Model Evaluation}
\begin{figure*}[htbp]
    \centering
    \vspace{-1mm}
    \includegraphics[width=0.95\linewidth]{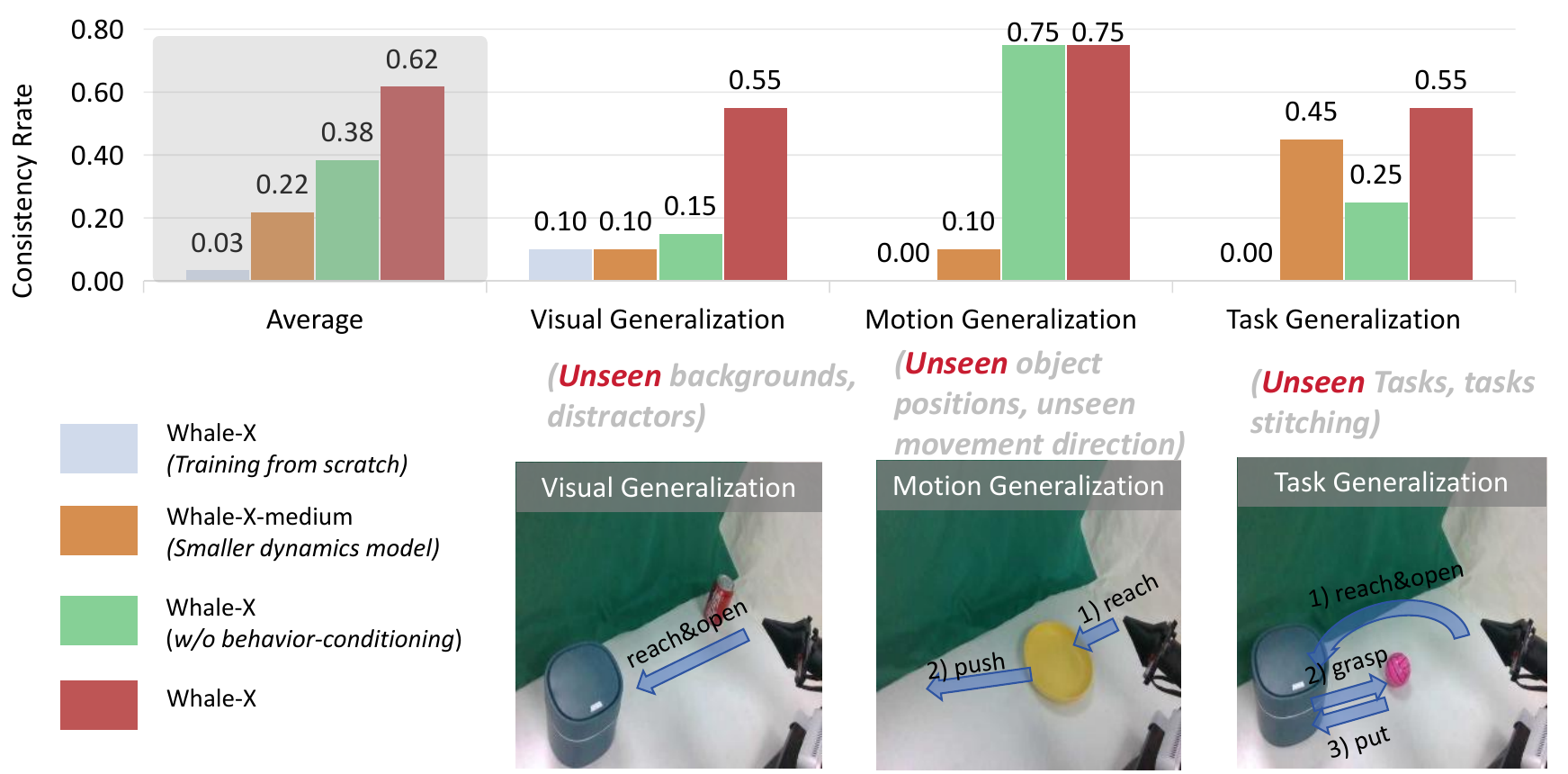}
    \caption{Physical robot evaluation on unseen scenarios. The row above shows the bar chart of the consistency rate, and the row below represents the tasks used for testing. The experiments demonstrate that Whale-X exhibits good generalization performance in unseen scenarios.}
    \vspace{-3mm}
    \label{fig:fig_plots_hist}
\end{figure*}

\textbf{Evaluation Metrics.} For a visual world model to be effective in decision-making, the world model needs to focus more on reasoning about the consequences of actions than on reconstructing irrelevant visual information like backgrounds. Thus we introduce the \textit{consistency rate} to assess whether the differences in reconstructed object positions, interactive object states, and robot arm positions fall within an acceptable range compared to the ground truth. We use the multimodal large model GPT-4o~\citep{achiam2023gpt} for this evaluation through multiple rounds of Q\&A. Details of the prompts and the evaluation process can be found in Appendix~\ref{app:GPT-4o evaluation}, with results presented in Figure~\ref{fig:fig_plots_hist}. In addition, we employ several video fidelity metrics, similar to those in Section~\ref{sec:exp_sim}, to assess the quality of video generation by the world models.
 
\textbf{Task Results}

\begin{table}[htbp]
    \centering
    \begin{tabular}{lccc}
    \toprule
    \textbf{Real-world Tasks} & PSNR$\uparrow$ & SSIM$\uparrow$ & LPIPS$\downarrow$ \\
    \midrule
    \multicolumn{4}{c}{\textit{unseen tasks \& 256$\times$256 resolution}} \\
    \midrule
    Whale-X-base (training from scratch)& 20.0 &74.9 & 37.0 \\
    Whale-X-base (w/o behavior-conditioning) & 21.4 & 79.0 & 31.2\\
    Whale-X-medium  (77M parameter) & 21.9 &79.9& 30.0\\
    Whale-X (ours) & \textbf{22.3} & \textbf{80.5} & \textbf{29.6}\\
    \bottomrule
    \end{tabular}
    \caption{Video fidelity of Whale-X on real-world tasks.}
    \vspace{-3mm}
    \label{tab:real_world_video}
\end{table}

 Whale-X shows a clear advantage in our real-world experiments. Specifically, as shown in Figure~\ref{fig:fig_plots_hist}, the quantitative results indicate that: 1) Whale-X improves consistency by \textbf{63\%} compared to models without behavior-conditioning, demonstrating that these mechanisms significantly enhance the OOD generalizability; and 2) Whale-X, pre-trained on 970k samples, achieved much higher consistency rate than models trained from scratch, highlighting the benefits of pre-training on large-scale internet data. 3) Increasing the model parameters improves the generalizability of the world model. Whale-X-base with 203M parameters dynamics model achieves a consistency rate that is about three times higher than the 77M version across three unseen tasks. Furthermore, the evaluation of video generation quality aligns with these consistency rate findings as illustrated in Table~\ref{tab:real_world_video}. The behavior-conditioning technique, pre-trained on large-scale datasets, combined with scaling up model parameters, significantly enhances OOD generalizability from a video fidelity perspective.

\subsubsection{Uncertainty Estimation} 
\textbf{Evaluation Metrics.} In contrast to simulated tasks, obtaining a pixel-based reward function in real-world tasks is nearly impossible, which makes it challenging to optimize policies directly through reinforcement learning within the world model.To overcome this limitation, we calculate cumulative uncertainty for each generated trajectory using retracing rollout and then select the top $k\%$ of trajectories with the lowest cumulative uncertainty. For these selected trajectories, we compute the consistency rate as an evaluation metric. Intuitively, if uncertainty accurately measures model error, lower uncertainty should correspond to more realistic generated trajectories, which are expected to yield a higher consistency rate.

\begin{figure*}[htbp]
    \centering
 \vspace{-2mm}
    \includegraphics[width=0.5\linewidth]{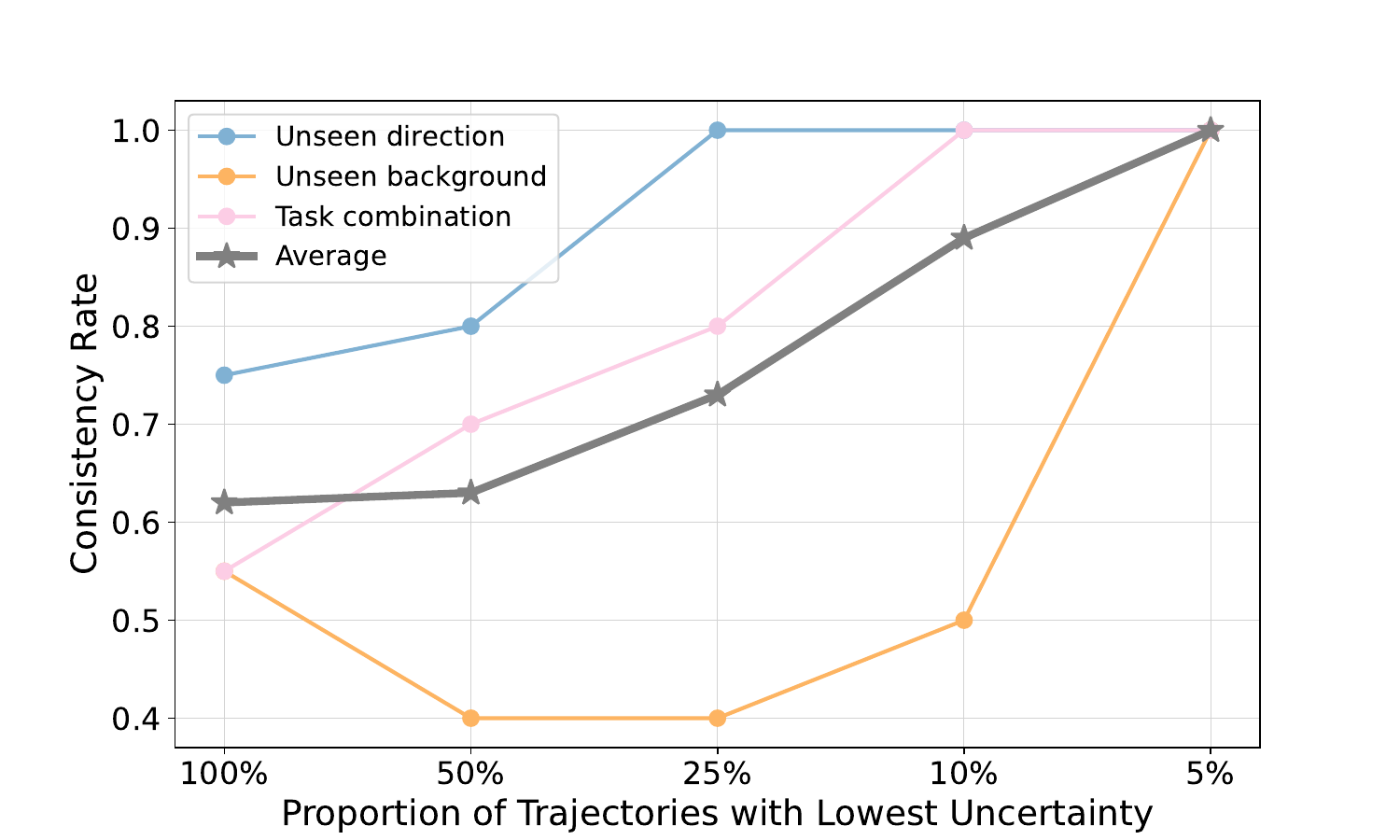}
	\caption{Consistency rate with different proportions of trajectories with the lowest uncertainty.} 
 \vspace{-3mm}
	\label{fig:uncertainty_real}
\end{figure*}

\textbf{Task Results.} Figure ~\ref{fig:uncertainty_real} presents the results of uncertainty estimation in real-world tasks, illustrating how the filtered proportion of trajectories affects the consistency rate. The gray line represents the average consistency rate across three tasks. The results show that reducing the number of selected trajectories—thus lowering the uncertainty threshold—leads to a marked increase in consistency rate. Notably, when $k=5$, the filtered trajectories align with real outcomes with $100\%$ accuracy. We observe a slight decrease in the consistency rate for visual generalization tasks at $k=10, 25,$ and $50$. In these cases, the estimated uncertainty tends to reflect model error on unseen backgrounds more than task-specific information, which could be one of the improvements for future work.

\subsection{Scaling Experiments}

In this section, we aim to investigate the scaling behavior of Whale-X. Specifically, We freeze the video tokenizer and behavior-conditioning model, adjusting only the model size and pre-training data size of dynamics models, considering the impact of model size and data size for the pre-training and fine-tuning phases. 

\begin{figure*}[htbp]
    \centering
\includegraphics[width=1.0\linewidth]{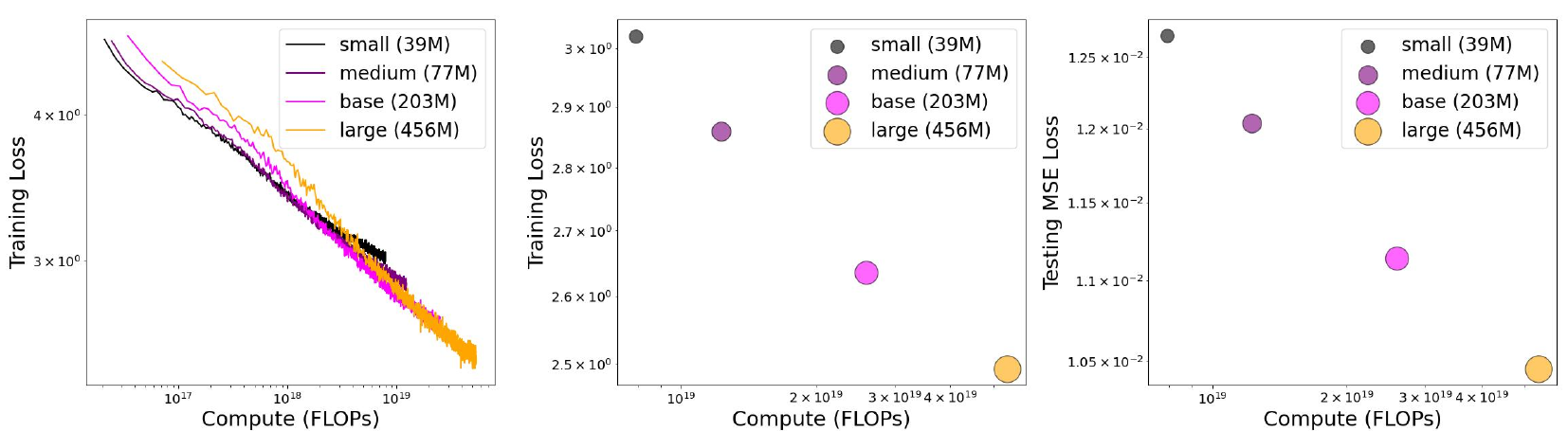}
    \caption{Scaling experiment results of Whale-X. The leftmost plot shows the training loss curves for models with varying parameter sizes during the pre-training phase. The second plot presents the final training loss for all models after 300k pre-training steps. The third plot displays the test loss after fine-tuning. The legend in the figure indicates the parameter number of the dynamics model. }
    \label{fig:scaling}
\end{figure*}

\textbf{Pre-training Scaling Experiments.} 
With a frozen video tokenizer and behavior-conditioning model, we train four dynamics models ranging from 39M to 456M parameters during the pre-training phase, with results shown in the first two plots of Figure \ref{fig:scaling}. These results demonstrate that Whale-X exhibits strong scalability, as increasing either the pre-training data or the number of parameters reduces the training loss. Notably, the training loss of Whale-X follows a log-linear relationship with FLOPs, which can guide the design of larger models and appropriate data ratios for future experiments.

\textbf{Fine-tuning Scaling Experiments.}
Apart from the scalability in the pre-training stage, it is also worth verifying whether a larger model can exhibit better performance during the fine-tuning phase. To this end, we fine-tune a series of dynamics models and show the test mean-squared-error losses in the leftmost plot in Figure \ref{fig:scaling}. The results indicate that after fine-tuning, the larger model demonstrates lower loss on test data, highlighting the promising scalability of Whale-X for real-world tasks.
\label{sec:exp_scale}

\subsection{Visualization Results}
\textbf{Qualitative Evaluation.} Figure \ref{fig:qualitative evaluation} demonstrates the qualitative evaluation results on Meta-World, Open X-Embodiment, and our real-world tasks. The results show that Whale-ST and Whale-X can generate high-fidelity video trajectories for long-horizon rollouts, maintaining visual quality and consistency over extended sequences. More qualitative evaluation results are provided in Appendix \ref{app:qualitative_eval}.

\begin{figure*}[!t]
    \centering
\includegraphics[width=0.95\linewidth]{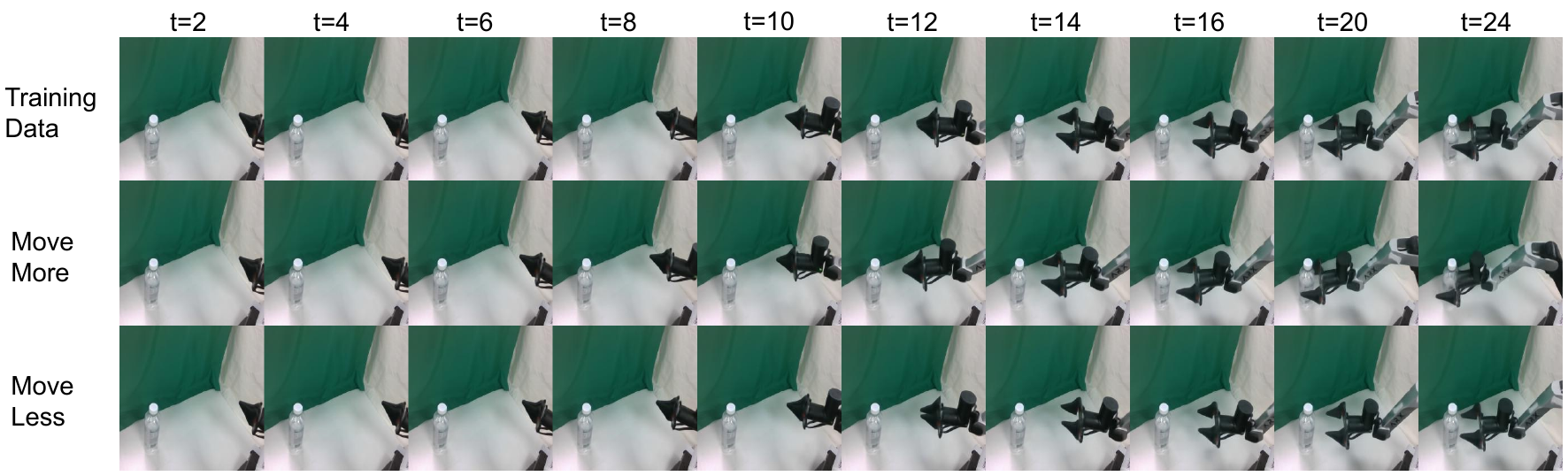}
    \caption{Controllable generation of Whale-X. Given an action sequence $a_{0:t}$, the first row shows the generated trajectory by interactively executing $a_{0:t}$ in Whale-X. The second row presents the "move more" trajectory generated by executing $1.2\cdot a_{0:t}$, and the third row shows the "move less" trajectory generated by executing $0.8\cdot a_{0:t}$.}
    \vspace{-2mm}
    \label{fig:controllable}
\end{figure*}

\textbf{Controllable Generation.} Figure \ref{fig:controllable} demonstrates the strong controllability and generalizability of Whale-X. Given an unseen action sequence, Whale-X can generate videos that align with human understanding, learning the causality between actions and robotic arm displacement.

\textbf{Behavior-conditioning Visualization.} To verify whether the learned behavior embedding has captured policy modes, we perform t-SNE~\citep{van2008visualizing} to visualize the representations corresponding to different tasks and policies. Figure \ref{fig:same_task_diff_policies} shows that different policies for the same task can be distinguished by the learned behavior embedding. Notably, the embedding of the noisy expert policy appears to be a linear interpolation between the expert policy and the noisy policy, indicating that the behavior-conditioning models the policies reasonably. Figure \ref{fig:diff_tasks_expert_policies} shows that the expert policies for different tasks can also be distinguished, while Figure \ref{fig:diff_tasks_rnd_policies} shows the random policies for different tasks cannot. This distinction indicates that our learned embedding is more inclined toward policy representation rather than task representation.

\section{Discussions and Limitations}
In this paper, We introduce WHALE, a framework of world model learning that incorporates the behavior-conditioning mechanism and retracing-rollout technique to enhance out-of-distribution generalization and efficient uncertainty estimation. Building on this foundation, we present Whale-ST, a scalable ST-transformer-based world model, and pre-train a 414M-parameter Whale-X on large-scale real-world robot data to assist physical robot manipulation. As a powerful world model with strong generalizability and promising scalability, Whale-X enables high-fidelity imagination and accurate value estimation, even in novel scenarios, thereby facilitating downstream control tasks.

\textbf{Limitations and future work.}\quad  Although WHALE marks significant progress, there remains substantial room for further improvement in future work. One limitation is the lack of diversity in real-world robotic data, typically collected by a narrow range of policies (e.g. near-optimal policies). This poses significant challenges to the generalization of world models. Additionally, we found that the quality of reward models with visual input plays a crucial role in accurate value estimation, which remains an unsolved challenge for future research. Lastly, we have to mention that although the generalization capability of Whale-ST and Whale-X has significantly improved compared with previous methods, it remains limited for zero-shot transfer in the face of the diversity and complexity of unseen real-world tasks. Integrating existing prior knowledge into the data-driven world model learning process could enable broader generalization, presenting a valuable avenue for long-term research.

\bibliographystyle{unsrt}  
\bibliography{reference}  

\newpage

\appendix
\section{Analysis of behavior-conditioning}
\label{app:analysis}

In this section, we provide some theoretical explanations about why behavior-conditioning mechanism helps mitigate the generalization error caused by the policy divergence. The analysis is mainly adapted from \cite{pcmodel}.

First, we introduce an assumption on the smoothness of a well-trained dynamics model:

\begin{assumption}
\label{app:assump}
    For the learned dynamics model $T$, the point-wise total-variation model error $D_{\mathrm{TV}}[T^*(\cdot|\tau_h), T(\cdot|\tau_h)]$ is $L$-Lipschitz with respect to the trajectory inputs, i.e.,
    $$
        \Big| D_{\mathrm{TV}}[T^*(\cdot|\tau^1_{h}),T(\cdot|\tau^1_{h})] - D_{\mathrm{TV}}[T^*(\cdot|\tau^2_{h}),T(\cdot|\tau^2_{h})]\Big|
        \leq L\cdot D(\tau^1_{h}, \tau^2_{h}) ,
    $$
    where $D(\cdot,\cdot)$ is some kind of distance defined on the trajectory space.
\end{assumption}

Assumption \ref{app:assump} measures the local extrapolation ability of a world model. Based on this assumption, the value gaps of common dynamics model $T$ without a behavior-conditioning mechanism can be controlled:

\begin{proposition}
\label{prop:pam}
    Under Assumption \ref{app:assump}, for any policy $\pi$, the value gap of common dynamics model $T$ without behavior-conditioning has an upper bound:
    $$
        \Big|V^\pi_T - V^\pi_{T^*}\Big| \leq 2R_{\max}H^2\Big(\underbrace{\sqrt{2~l_\mathrm{KL}(T;\Pi)}}_{\mathrm{in~distribution~error}}+\underbrace{L\cdot W_1(d^{\pi},d^{\Pi})}_{\textcolor{red}{\mathrm{policy~divergence}}}\Big),
        \label{app:pam}
    $$
    where $W_1(d^\pi,d^\Pi)$ is the Wasserstein-1 distance between the $\pi$-induced trajectory distribution $d^\pi(\tau)$ and the behavior trajectory distribution $d^\Pi(\tau)=\mathbb E_{\mu\sim\Pi}[d^\mu(\tau)]$.
\end{proposition}

Proposition \ref{prop:pam} shows that the generalization of common dynamics model $T$ solely relies on its point-level smoothness over the trajectory inputs, resulting in an inevitable extrapolation error of the policy distribution. In contrast, a policy-conditioned dynamics model $T(\cdot)$, which yields adapted dynamics model $T(\pi)$ for some policy $\pi$, takes a further step to reduce the policy distribution extrapolation error:

\begin{proposition}
\label{prop:pcm}
    Under Assumption \ref{app:assump}, for any policy $\pi$, the value gap of policy-conditioned dynamics model $T(\cdot)$  has an upper bound:
    $$
        \Big|V^\pi_{T(\pi)} - V^\pi_{T^*}\Big| \leq 2R_{\max}H^2\Big(\underbrace{\sqrt{2~l_\mathrm{KL}(T;\Pi)}}_{\mathrm{in~distribution~error}}+\underbrace{L\cdot W_1(d^{\pi},d^{\Pi})-C(\pi,\Pi)}_{\textcolor{red}{\mathrm{reduced~policy~divergence}}}\Big),
        \label{app:pcm}
    $$
    where the adaptation gain $\displaystyle C(\textcolor{blue}{\pi},\textcolor{red}{\Pi}):=\mathbb E_{\textcolor{red}{\mu\sim\Pi}}\mathbb E_{\tau\sim \textcolor{blue}{d^{\pi}}}D_{\mathrm{TV}}[T^*,T(\textcolor{red}{\mu})](\tau) - \mathbb E_{\tau\sim \textcolor{blue}{d^{\pi}}} D_{\mathrm{TV}}[T^*,T(\textcolor{blue}{\pi})](\tau)$ summarizes the policy adaptation effect.
\end{proposition}

Proposition \ref{prop:pcm} explains the benefit brought by behavior-conditioning: a positive adaptation gain $C(\pi,\Pi)$, which quantifies the advantage of the policy adaptation effect. The key insight is that when testing on an unseen policy $\pi$ within some effective region, the model $T(\pi)$, customized for $\pi$, should exhibit a smaller model error under the target trajectory distribution $d^\pi$ compared to models $T(\mu)$ trained on behavior policies $\mu\in\Pi$, which mitigates the generalization error caused by the policy extrapolation. Although it is challenging to rigorously analyze the adaptation gain $C(\pi,\Pi)$ due to the complexity of neural networks and the optimization process, qualitative discussions and empirical evidence, as shown in \cite{pcmodel}, justify the underlying rationale.

\section{Implementation Details}
\subsection{Implementation Details of Whale-ST and Whale-X}
\label{impl:whale}
\paragraph{Video Tokenizer.} Here we show the architecture and training hyperparameter of the video tokenizer as shown in Table~\ref{tab:tokenizer_hyperparameters}. We train three different video tokenizers in total.
\begin{table}[htbp]
    \centering
    \begin{tabular}{l|cccc}
        \toprule
        \textbf{Component} & \textbf{Parameter} & \textbf{Whale-ST}$_{(64\times64)}$ & \textbf{Whale-ST}$_{(256\times256)}$ & \textbf{Whale-X}$_{(256\times256)}$ \\ \midrule
        \multirow{3}{*}{Encoder} & num\_layers & 4 & 12 & 12 \\ 
                                 & d\_model & 512 & 512 & 512 \\
                                 & num\_heads & 8 & 8 & 8 \\ \midrule
        \multirow{3}{*}{Decoder} & num\_layers & 8 & 16 & 20 \\ 
                                 & d\_model & 512 & 512 & 1024 \\
                                 & num\_heads & 8 & 8 & 16 \\\midrule
        \multirow{4}{*}{Codebook} & num\_codes & 1024 & 1024 & 2048 \\ 
                                  & patch\_size & 4 & 16 & 16 \\
                                  & latent\_dim & 32 & 32 & 32 \\
                                  & beta & 0.25 & 0.25 & 0.25 \\ 
                                  \midrule
        \multirow{9}{*}{Optimizer} & type & AdamW &AdamW & AdamW  \\
                                & max\_lr & 3e-4 &3e-4 & 3e-4 \\
                                & min\_lr & 3e-4 & 3e-4 & 3e-5 \\
                                & $\beta_1$ & 0.9 & 0.9 & 0.9 \\
                                & $\beta_2$ & 0.9 & 0.9 & 0.9 \\
                                & weight\_decay & 1e-4 & 1e-4 & 0 \\
                                & warmup\_steps & 10k & 10k & 5k \\
                                & batch\_size & 32 & 32 & 64 \\ 
                                & training\_steps & 100k & 150k & 300k \\ 
                                \bottomrule
    \end{tabular}
    \caption{Hyperparameter of video tokenizers.}
    \label{tab:tokenizer_hyperparameters}
\end{table}

\paragraph{Behavior-conditioning Model.} The model architecture and training hyperparameters of the behavior-conditioning model are shown in Table~\ref{tab:policy_embed_hyperparameter}. We also train three different behavior embedding models for Whale-ST and Whale-X. Additionally, We also observe overfitting in the behavior-conditioning model during pre-training, prompting the use of the early-stop technique. As a result, the checkpoint at 50k is selected as the final model for Whale-X.

\begin{table}[htbp]
    \centering
    \begin{tabular}{l|cccc}
        \toprule
        \textbf{Component} & \textbf{Parameter} & \textbf{Whale-ST}$_{(64\times64)}$ & \textbf{Whale-ST}$_{(256\times256)}$ & \textbf{Whale-X}$_{(256\times256)}$ \\ \midrule
        \multirow{4}{*}{Posterior} & num\_layers & 8 & 8 &12 \\ 
                                 & d\_model & 512 & 512 & 768\\
                                 & num\_heads & 8 & 8 &12\\ 
                                 &patch\_size & 8 & 32&32 \\ \midrule
        \multirow{4}{*}{Prior} & num\_layers & 4 & 4  & 8\\ 
                                 & d\_model & 512 & 512 &512 \\
                                 & num\_heads & 4 & 4 & 8\\ 
                                 &patch\_size & 8 & 32&32 \\ \midrule
        \multirow{4}{*}{Policy} & num\_layers & 8 & 8 &12 \\ 
                                 & d\_model & 512 & 512& 768\\
                                 & num\_heads & 8 & 8 &12\\ 
                                 & log\_std & [-2, 5] & [-2, 5]&[-2, 5] \\ 
                                 &patch\_size & 8 & 32&32 \\ 
                                 \midrule
        \multirow{2}{*}{Embedding} & category\_size & 16 & 16&16 \\ 
                                 & class\_size & 16 & 16 &16\\
                                 \midrule
        \multirow{10}{*}{Optimizer} 
                                & type & AdamW &AdamW & AdamW  \\
                                & max\_lr & 3e-4 &3e-4 & 3e-4  \\
                                & min\_lr & 3e-5 & 3e-5 & 3e-5\\
                                & $\beta_1$ & 0.9 & 0.9  & 0.9\\
                                & $\beta_2$ & 0.9 & 0.9 & 0.9\\
                                & weight\_decay & 1e-4 & 1e-4& 1e-4\\
                                & warmup\_steps & 5k & 5k &5k\\
                                & batch\_size & 64 &  64 &64\\ 
                                & training\_steps & 100k & 100k& 50k \\
                                \bottomrule
    \end{tabular}
    \caption{Hyperparameter of behavior-conditoning models.}
    \label{tab:policy_embed_hyperparameter}
\end{table}

\paragraph{Dynamics model} Table~\ref{tab:dynamic_hyperparameter} and Table~\ref{tab:dynamic_hyperparameter2} present the hyperparameters of the dynamics model. We train a total of 6 different dynamics models. The architecture design and training hyperparameters of our dynamics model are also referred to \cite{bruce2024genie}.

\begin{table}[htbp]
    \centering
    \begin{tabular}{l|ccccc}
        \toprule
        \textbf{Model} & \textbf{\#Parameters (dynamics only)} & \textbf{num\_layers} & \textbf{num\_heads} & \textbf{d\_model} \\ 
        \midrule
        Whale-ST (64)  & 26M& 12 & 8  & 512  \\ 
        Whale-ST (256)  & 26M& 12 & 8 & 512  \\ 
        Whale-X-small & 39M& 18 & 8 & 512 \\ 
        Whale-X-medium & 77M& 16 & 16 & 768 \\ 
        Whale-X-base & 204M & 24 & 16& 1024 \\ 
        Whale-X-large & 456M & 24 & 12 & 1536 \\ 
        \bottomrule
    \end{tabular}
    \caption{Model hyperparameter of dynamics models.}
    \label{tab:dynamic_hyperparameter}
\end{table}

\begin{table}[htbp]
    \centering
    \begin{tabular}{c|c}
    \toprule
        \textbf{Parameter} & \textbf{Value} \\ \midrule
        max\_lr & 3e-5 \\ 
        min\_lr & 3e-6 \\ 
        $\beta_1$ & 0.9 \\ 
        $\beta_2$ & 0.9 \\ 
        weight\_decay & 0 \\ 
        warmup\_steps & 5k \\ 
        batch\_size & 64 \\ 
        training\_steps & 300k \\ \bottomrule
    \end{tabular}
    \caption{Trainig hyperparameter of dynamics models.}
    \label{tab:dynamic_hyperparameter2}
\end{table}

\subsection{Fine-tuning Details of Whale-X} 
\label{app:finetune}
For fine-tuning all pre-trained models, we first update the video tokenizer for 5000 gradient steps while keeping the encoder network fixed. After that, we update the behavior-conditioning model for 1000 gradient steps, and finally, we update the dynamics model for 5000 gradient steps. For training models from scratch, the video tokenizer, behavior-conditioning model, and dynamics model are all updated for 10,000 gradient steps.

\subsection{Implementation Details of Baselines}
\label{impl:baselines}
\paragraph{Baselines for model evaluation} We use the official implementation of VP2~\citep{vp2} for both FitVid and MCVD. For DreamerV3, we retain only the world model learning component. Additionally, we use the official implementation of iVideoGPT as described in their original paper, but with a reduced number of parameters. The detailed hyperparameters for DreamerV3 and iVideoGPT are provided in Table~\ref{hyper:dreamer} and Table~\ref{hyper:ivideo}, respectively.
\paragraph{Baselines for uncertainty estimation} We implement both entropy-based and ensemble-based uncertainty estimation methods based on Whale-ST. The entropy-based method directly uses the entropy of the logits output by the dynamics model as the measure of uncertainty. For the ensemble-based method, we independently train three dynamics models and use the pixel-level variance between their output images as the measure of uncertainty. 
\begin{table}[htbp]
    \centering
    \begin{minipage}[t]{0.45\linewidth}
      \vtop{
      \centering
      \begin{tabular}{c|c}
      \toprule
           \textbf{Hyperparameters} & \textbf{Values}  \\ \midrule
           \# Parameters & 44M  \\
           Dynamics hidden & 1024  \\
           Dynamics deterministic & 1024 \\
           Dynamics stochastic & 32 \\
           Dynamics discrete & 32 \\
           CNN depth & 64 \\
           CNN kernel size  & 4 \\
           MLP layers  & 5 \\
           MLP units  & 1024\\
           Actionvation & SiLU \\
           Train batch size & 32 \\
           Train batch length  & 8 \\
      \bottomrule
      \end{tabular}
      \caption{Hyperparameters for DreamerV3.}
      \label{hyper:dreamer}
      }
    \end{minipage}%
    \hfill
    \begin{minipage}[t]{0.45\linewidth}
      \vtop{
      \centering
      \begin{tabular}{c|c}
      \toprule
           \textbf{Hyperparameters} & \textbf{Values}  \\ \midrule
           \# Parameters & 63M  \\
           Down blocks & 3  \\
           Down layers per block & 2 \\
           Down channels & [64, 128, 256] \\
           Up blocks & 3 \\
           Up layers per block & 3 \\
           Up channels  & [256, 128, 64] \\
           Embedding dim  & 64 \\
           Codebook size  & 8192\\
           Actionvation & SiLU \\
           Transformer hidden dim & 512 \\
           Transformer hidden layers & 6 \\
           Attention Heads & 8 \\
           Feedforward dim  & 1024 \\
      \bottomrule
      \end{tabular}
      \caption{Hyperparameters for iVideoGPT.}
      \label{hyper:ivideo}
      }
    \end{minipage}
\end{table}

\subsection{Implementation Details of Evaluation Metrics}
\label{impl:metrics}
\paragraph{Evaluation metrics for model evaluation}

The metrics we use for model evaluation are defined as follows:

\textbf{Absolute Error} The absolute error is defined as the difference between the value and estimated value of a policy:
\begin{equation}
    \text{AbsErr} = | V^{\pi}-\hat{V}^{\pi} |,
\end{equation}
where $V^{\pi}$ is the true value of the policy and $\hat{V}^{\pi}$ is the estimated value of the policy.

\textbf{Rank correlation} Rank correlation measures the correlation between the ordinal rankings of the value estimates and the true values, which can be written as:
\begin{equation}
    \text{RankCorr}=\frac{\operatorname{Cov}(V_{1: N}^\pi, \hat{V}_{1: N}^\pi)}{\sigma(V_{1: N}^\pi) \sigma(\hat{V}_{1: N}^\pi)},
\end{equation}
where $1:N$ denotes the indices of the evaluated policies.

\textbf{Regret@k} Regret@k is the difference between the value of the best policy in the entire set, and the value of the best policy in the top-k set (where the top-k set is chosen by estimated values). It can be defined as:
\begin{equation}
    \text{Regret @} \mathrm{k}=\max _{i \in 1: N} V_i^\pi-\max _{j \in \operatorname{topk}(1: N)} V_j^\pi,
\end{equation}
where topk($1:N$) denotes the indices of the top K policies as measured by estimated values $\hat{V}^{\pi}$.

\paragraph{Evaluation metrics for uncertainty estimation} For model error prediction, we use unseen action sequences to rollout 8 steps in Whale-ST, and measure model error using the perceptual loss between the generated samples and the real samples. For offline MBRL evaluation, Figure \ref{fig:offline_rl} shows the success rate over gradient update steps of different uncertainty estimation algorithms across five different tasks. For each evaluation, we sample 20 trajectories to calculate the success rate of the current policy, with a maximum trajectory length truncated at 100. Additionally, we set the uncertainty penalty coefficients as follows: 50 for retracing rollout, 5000 for the ensemble-based method, and 0.1 for the entropy-based method. Additionally, we set the model rollout horizon to 4, the mix ratio of real and generated samples to 0.5, and kept the hyperparameters consistent across all tasks.

\section{Data Preparation}
\subsection{Simulated Data}
We select a total of 20 tasks from the MetaWorld benchmark. Each task includes a training set of 3,000 trajectories and a test set of 1,500 trajectories. Specifically, for each task, we use six different policies to collect the training set: expert policy, random policy, two suboptimal policies with different levels of Gaussian noise, and two cross-environment policies. Additionally, three unseen policies are used to gather the testing data. The world models are trained on the full training dataset, followed by a thorough evaluation using the testing data.

\label{data:sim}
\subsection{Pre-training Data}
\label{data:pretrain}
Follow~\cite{openvla}, our pre-training dataset collection includes 27 datasets, with a total scale of 970k demonstrations, as shown in Table~\ref{tab:data_mix}.


\begin{table}[th]
    \centering
       \begin{tabular}{lr}
        \toprule
        \textbf{Whale-X Pre-training Dataset Mixture} & \textbf{Percentage}\\
        \midrule
        \textbf{Fractal}~\citep{brohan2022rt} & 12.7\% \\
        Kuka~\citep{kalashnikov2018qt} & 12.7\% \\
        \textbf{Bridge}\citep{ebert2021bridge, walke2023bridgedata} & 13.3\% \\
        Taco Play~\citep{rosetebeas2022latent,mees2023grounding} & 3.0\% \\
        \textbf{Jaco Play}~\citep{dass2023jacoplay} & 0.4\% \\
        Berkeley Cable Routing~\citep{luo2023multistage} & 0.2\% \\
        \textbf{Roboturk}~\citep{DBLP:journals/corr/abs-1811-02790} & 2.3\% \\
        \textbf{Viola}~\citep{zhu2023viola} & 0.9\% \\
        Berkeley Autolab UR5~\citep{BerkeleyUR5Website} & 1.2\% \\
        \textbf{Toto}~\citep{zhou2023train} & 2.0\% \\
        \textbf{Language Table}~\citep{lynch2023interactive} & 4.4\% \\
        \textbf{Stanford Hydra Dataset}~\citep{belkhale2023hydra}  & 4.4\% \\
        Austin Buds Dataset~\citep{zhu2022bottom}  & 0.2\% \\
        NYU Franka Play Dataset~\citep{cui2022play}  & 0.8\% \\
        Furniture Bench Dataset~\citep{heo2023furniturebench}  & 2.4\% \\
        UCSD Kitchen Dataset~\citep{ucsd_kitchens}  & <0.1\% \\
        Austin Sailor Dataset~\citep{nasiriany2022sailor}  & 2.2\% \\
        Austin Sirius Dataset~\citep{liu2022robot}  & 1.7\% \\
        DLR EDAN Shared Control~\citep{quere_shared_2020}  & <0.1\% \\
        IAMLab CMU Pickup Insert~\citep{saxena2023multiresolution}  & 0.9\% \\
        \textbf{UTAustin Mutex}~\citep{shah2023mutex} & 2.2\% \\
        Berkeley Fanuc Manipulation~\citep{fanuc_manipulation2023} & 0.7\% \\
        CMU Stretch~\citep{mendonca2023structured} & 0.2\% \\
        \textbf{BC-Z}~\citep{jang2022bc} & 7.5\% \\
        FMB Dataset~\citep{luo2024fmb}  & 7.1\% \\
        DobbE~\citep{shafiullah2023dobbe}  & 1.4\% \\
        DROID~\citep{khazatsky2024droid}  & 10.0\% \\
                \bottomrule
        \end{tabular}%
        \caption{Whale-X Pre-training Dataset Mixture.}
        \label{tab:data_mix}
\end{table}

\section{Real-world Task Design}
\label{app:data collection}
\subsection{Hardware Setup}
\begin{figure*}[!t]
    \centering
    \vspace{-2mm}
    \includegraphics[width=0.95\linewidth]{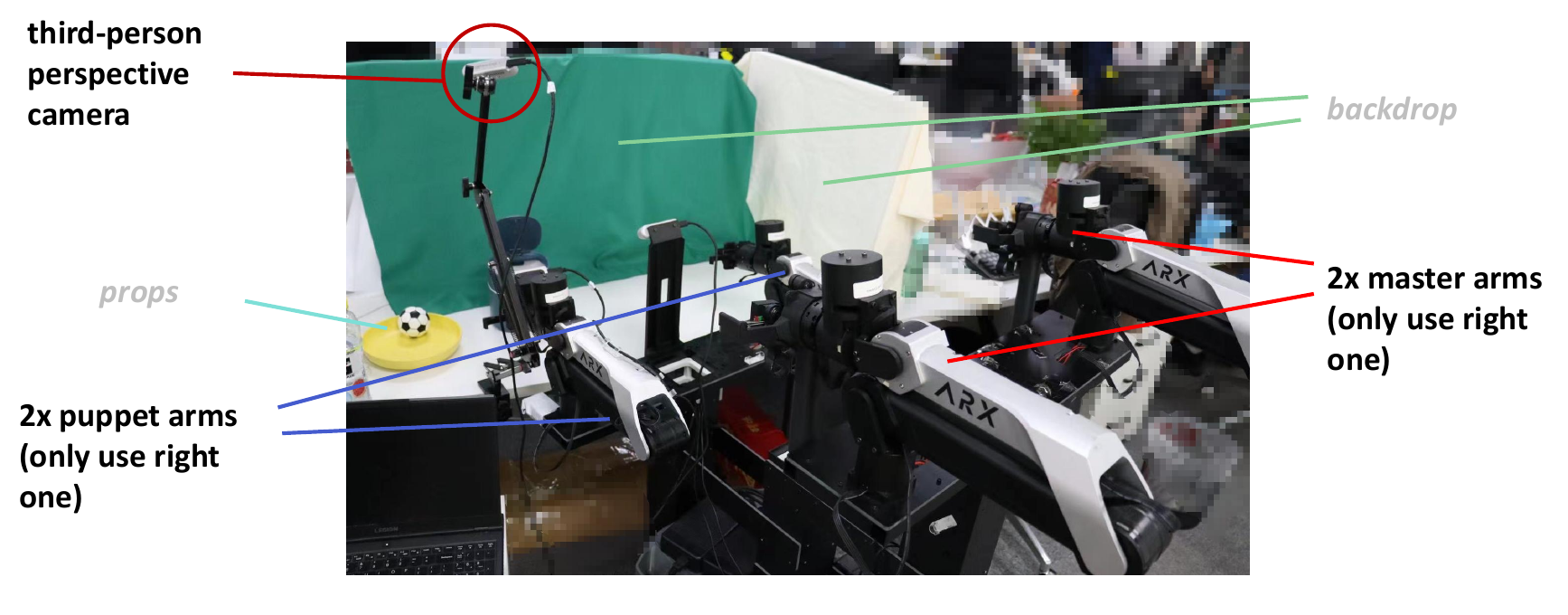}
    \vspace{-5mm}
    \caption{The illustration of our robotics platform used for physical robot evaluation.}
    \label{fig:fig_plots_hardware}
\end{figure*}

Our hardware setup is shown in Figure \ref{fig:fig_plots_hardware}. For the embodiment, we use the ARX5 robotic platform, which is similar to Aloha~\citep{mobile_aloha} and includes two master arms and two puppet arms. Data is collected via teleoperation and we only use the right arm in our experiment. For the vision sensor, a Realsense D435i camera is mounted above the left side of the platform to capture RGB image observations.

\subsubsection{Details of Tasks}
\label{exp:task_design}

The training data set used for finetuning consists of 4 tasks: \textbf{Move Bottle}, \textbf{Open Bin}, \textbf{Push Plate}, and \textbf{Throw Ball}. 

\textbf{Move Bottle}: The robot arm must first grasp the bottle and then move it to a specific area on the right side of the table. The bottle's initial position is somewhat random, varying within a range of two bottle widths around the location shown in the figure, while its target position remains fixed.
\begin{figure*}[!t]
    \centering
    \vspace{-2mm}
    
    \begin{minipage}[b]{0.43\linewidth}
        \centering
        \includegraphics[width=\linewidth]{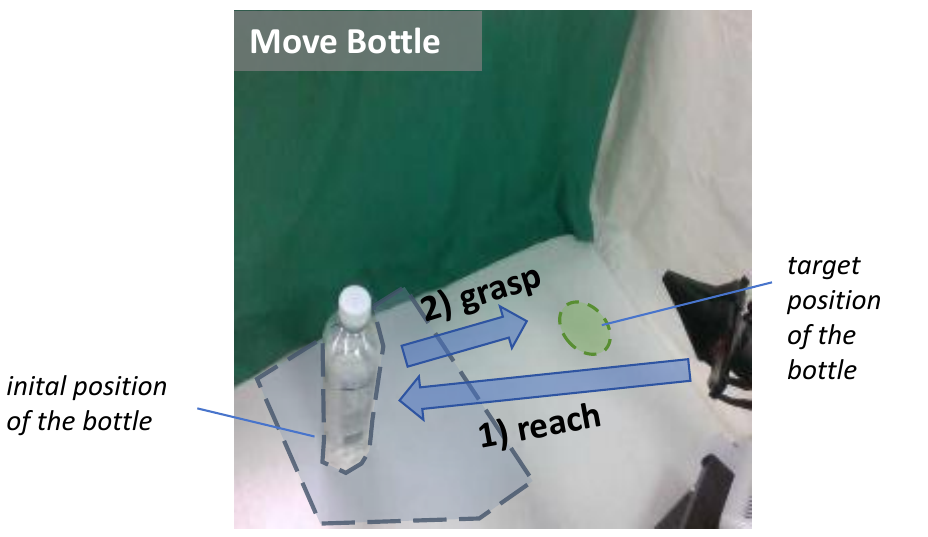}
        \subcaption{Move Bottle}
        \label{fig:fig_plots_training_move_bottle}
    \end{minipage}
    \begin{minipage}[b]{0.42\linewidth}
        \centering
        \includegraphics[width=\linewidth]{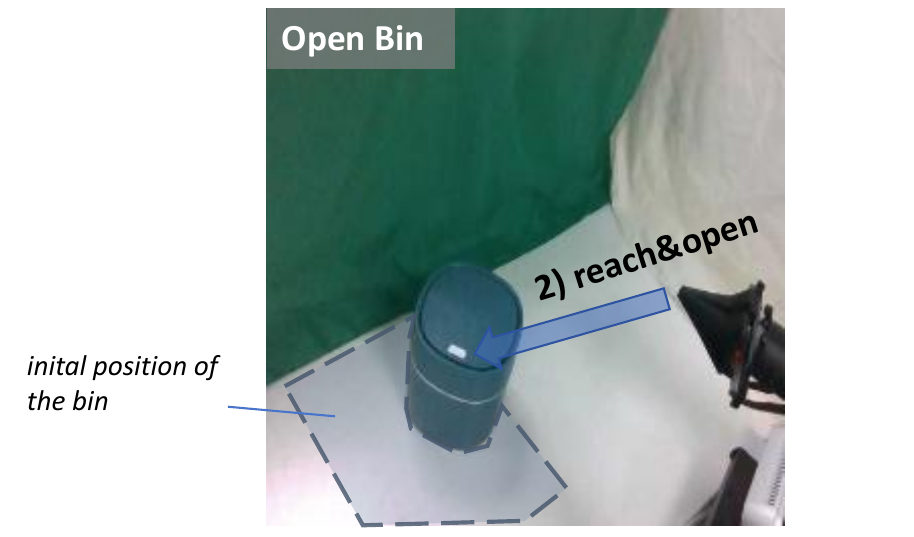}
        \subcaption{Open Bin}
        \label{fig:fig_plots_training_open_bin}
    \end{minipage}
    
    \vspace{4mm}
    
    \begin{minipage}[b]{0.48\linewidth}
        \centering
        \includegraphics[width=\linewidth]{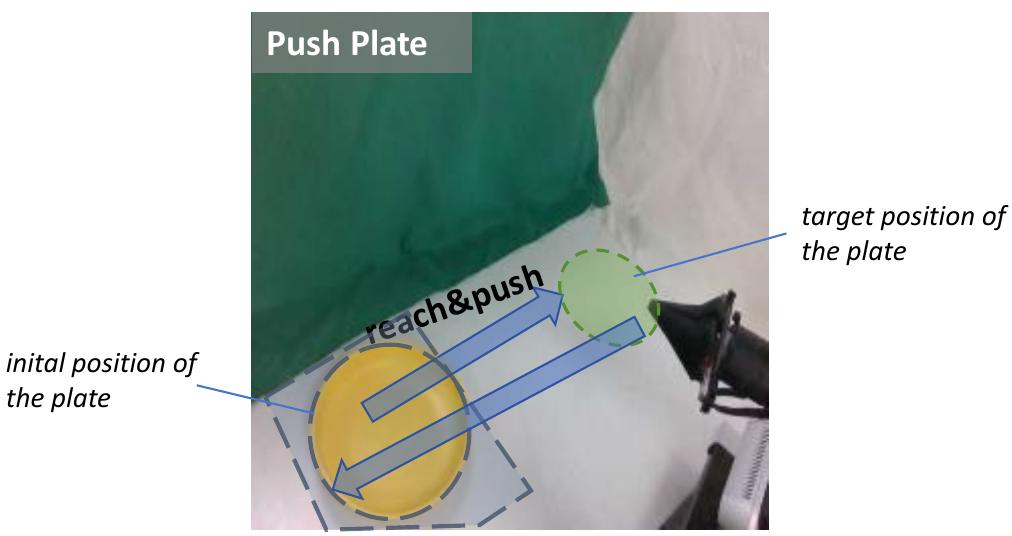}
        \subcaption{Push Plate}
        \label{fig:fig_plots_training_push_plate}
    \end{minipage}
    \begin{minipage}[b]{0.48\linewidth}
        \centering
        \includegraphics[width=\linewidth]{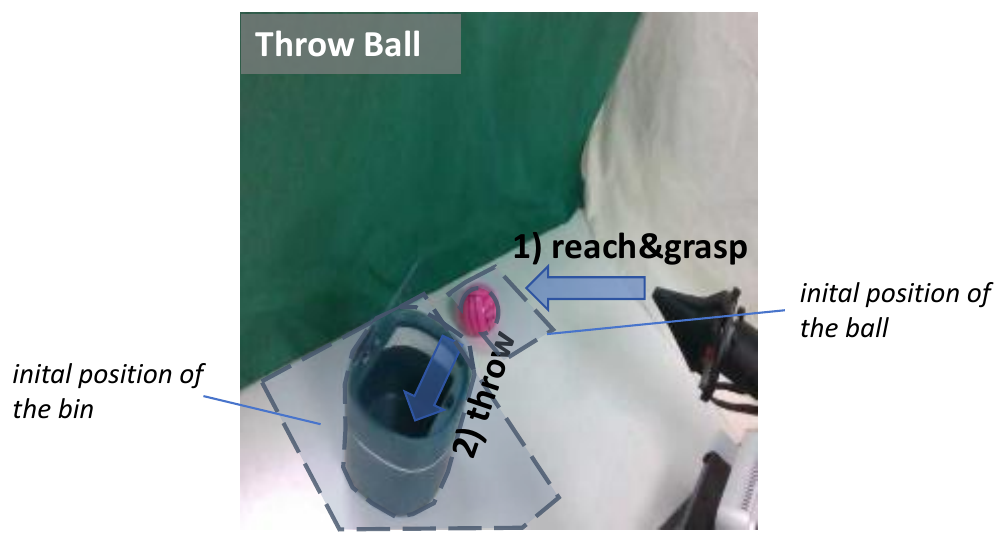}
        \subcaption{Throw Ball}
        \label{fig:fig_plots_training_throw_ball}
    \end{minipage}
    
    \vspace{-2mm}
    \caption{The illustration of Training Tasks.}
    \label{fig:training_tasks}
\end{figure*}

\textbf{Open Bin}: In this task, the robot arm must press a small white area on the trash bin's lid to open it. The initial position of the trash bin has some randomness, and it may vary within a 5 cm range around the position shown in the figure. Additionally, the orientation of the trash bin may have a random variation of about 10 degrees relative to its square alignment.

\textbf{Push Plate}: In this task, the robot arm must push the plate from the left side of the table to the right with appropriate force and angle. The challenge lies in the fact that the plate may rotate or shift during the pushing process. The initial position of the plate has some randomness, varying within a 5 cm range around the position shown in the figure. The robot arm needs to push the plate at a distance of approximately 20 cm.

\textbf{Throw Ball}: In this task, the robot arm needs to make a two-stage decision: 1) move to the ball's location and grasp it; 2) move to the trash bin's opening and release the gripper. The initial position of the trash bin has some randomness, varying within a 5 cm range around the position shown in the figure. 

\begin{figure*}[!t]
    \centering
    \vspace{-2mm}
    
    \begin{minipage}[b]{0.32\linewidth}
        \centering
        \includegraphics[width=\linewidth]{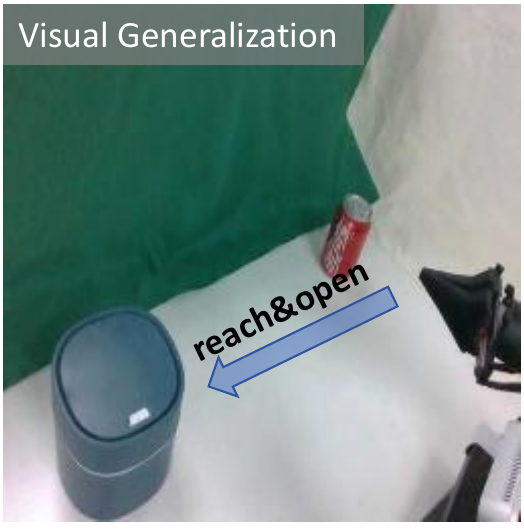}
        \subcaption{Visual Generalization}
        \label{fig:fig_plots_source_overview_vg}
    \end{minipage}
    \begin{minipage}[b]{0.32\linewidth}
        \centering
        \includegraphics[width=\linewidth]{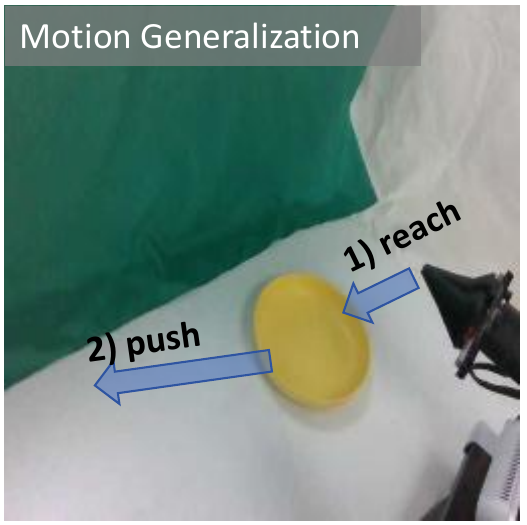}
        \subcaption{Motion Generalization}
        \label{fig:fig_plots_source_overview_mg1}
    \end{minipage}
    \begin{minipage}[b]{0.32\linewidth}
        \centering
        \includegraphics[width=\linewidth]{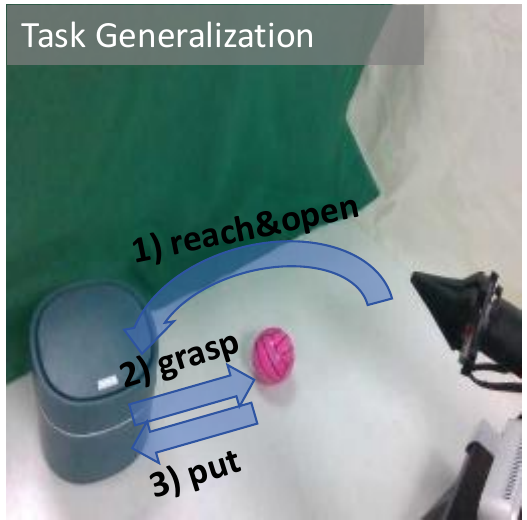}
        \subcaption{Task Generalization}
        \label{fig:fig_plots_source_overview_tg}
    \end{minipage}
    
    \vspace{-2mm}
    \caption{The illustration of Generalization Tasks.}
    \label{fig:generalization_tasks}
\end{figure*}

\textbf{Visual Generalization}: In this unseen scenario, we introduced several visual distractors not encountered during the fine-tuning phase, including a soda can, a plate, a ball, and a pencil, based on the Open Bin task. This task is designed to test the robustness of the world model's visual representation and its generalization in visual perception.

\textbf{Motion Generalization}: In this unseen scenario, based on the Push Plate task, we changed the specific task from pushing left to right in the fine-tuning data to pushing right to left. This task is designed to evaluate the model's ability to generalize environment transition modeling when facing an unseen action distribution, or even a completely reversed action distribution.

\textbf{Task Generalization}: In this unseen scenario, we combined two tasks from the fine-tuning phase—Open Bin and Throw Ball—into a new two-stage task. In this task, the robot arm must first open the bin and then place the ball inside. This task is designed to test the model's generalization ability to new tasks, as well as its capability to model long-horizon actions.

\subsection{Data Overview}
\begin{table}[h]
\centering
\begin{tabular}{c|c}
\toprule
\textbf{Entry} & \textbf{Value}\\
\toprule
\# Episodes & 300(240 for fine-tuning, 60 for testing) \\
Average horizon & 30 \\
Data Collect Method & Human teleoperation using the master arm \\
Scene Type & Table top \\
Robot Morphology & Single arm \\
Camera resolution & 640x480 \\
\# Cameras & 1 \\
Action dimension & 7 \\
Action space & EEF position \\
Action semantics & ($\Delta x$, $\Delta y$, $\Delta z$, $\Delta$roll, $\Delta$pitch, $\Delta$yaw, the gripper state) \\
Control frequency & 5Hz \\
Has suboptimal? & Yes(some failure data for fine-tuning) \\
Has camera calibration? & No \\
\bottomrule
\end{tabular}
\label{tab:collect_meta_info}
\caption{The meta Information of data used in physical robot evaluation.}
\end{table}

\section{Additional Experimental Results}

\subsection{Qualitative Evaluation}
\label{app:qualitative_eval}
\begin{figure*}[h!]
    \centering
    \includegraphics[width=1.0\linewidth]{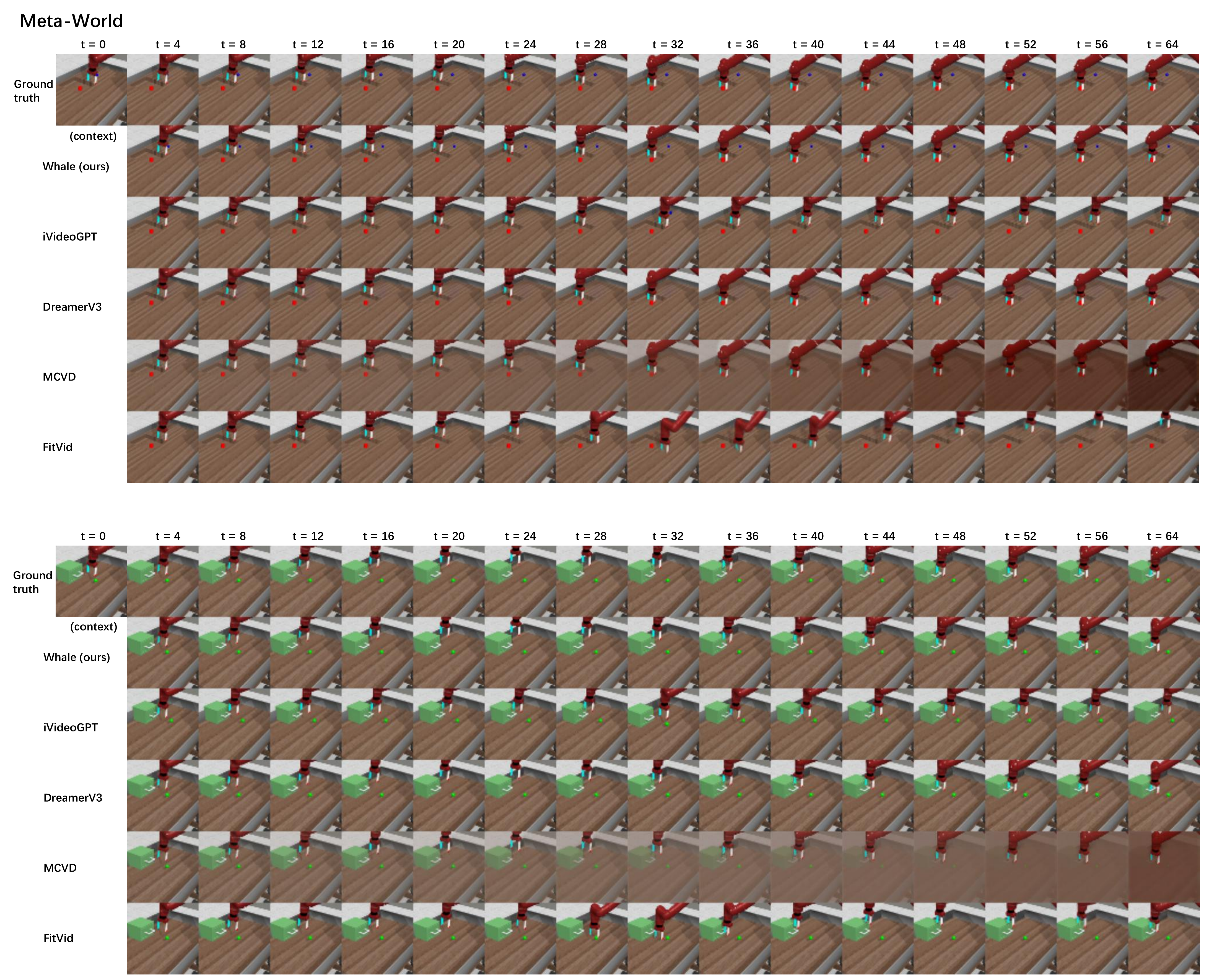}
    \caption{Additional qualitative evaluation on the Meta-World dataset. }
    \label{fig:qualitative_metaworld}
\end{figure*}

\paragraph{Qualitative Evaluation on Simulated Task}
Figure~\ref{fig:qualitative_metaworld} shows the results of Whale-ST and baselines after rolling out 64 steps in two different tasks. Notably, this qualitative evaluation is highly challenging and presents significant complexities. First, the evaluation rollout horizon is set to 64, exceeding that used in prior works, which imposes substantial demands on the generalizability and robustness of world models. Moreover, the variations between adjacent frames are subtle in the Meta-World environment, requiring world models to learn the semantics of actions from these minimal changes. In each image, the first row represents the real trajectory, while the others show the generated trajectories. It can be observed that Whale-ST not only generates high-fidelity videos but also accurately restores the robot arm's pose. DreamerV3 is the baseline closest to Whale, but its generated trajectory still loses key information, such as the blue marker representing the target point. The other baselines fail to accurately model the robot arm's pose changes from the subtle variations between adjacent frames.

\begin{figure*}[h!]
    \centering
    \includegraphics[width=1.0\linewidth]{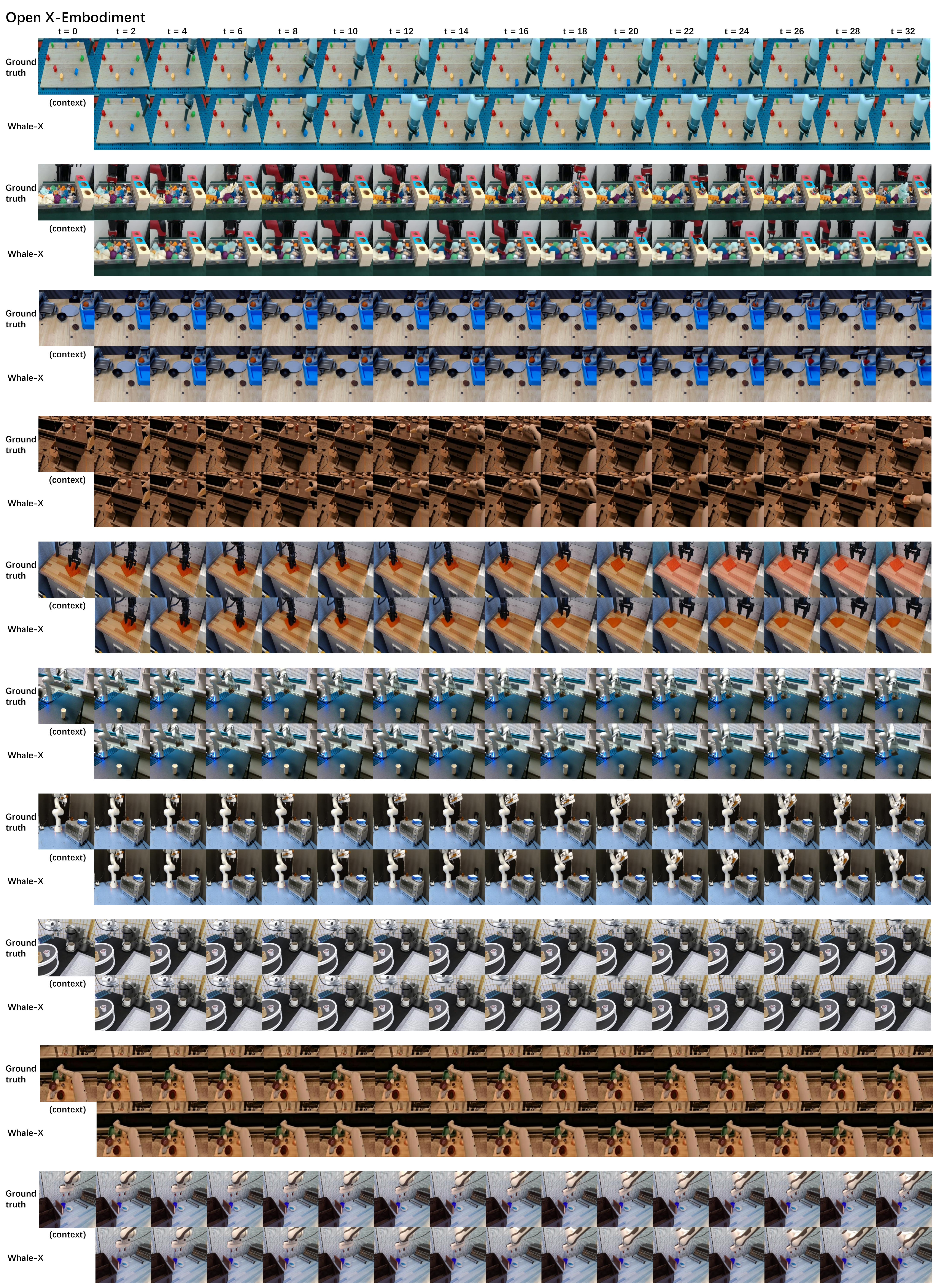}
    \caption{Additional qualitative evaluation on the Open X-Embodiment dataset. }
    \label{fig:qualitative_oxe}
\end{figure*}

\paragraph{Qualitative Evaluation on Open X-Embodiment Dataset} Figure~\ref{fig:qualitative_oxe} shows the qualitative evaluation results of Whale-X on Open X-Embodiment dataset. Whale-X demonstrates a remarkable ability to generate high-fidelity, action-conditioned trajectories. Moreover, with the aid of retracing-rollout and behavior-conditioning techniques, Whale-X consistently delivers highly accurate predictions of the robotic arm's pose.

\begin{figure*}[h!]
    \centering
    \includegraphics[width=1.0\linewidth]{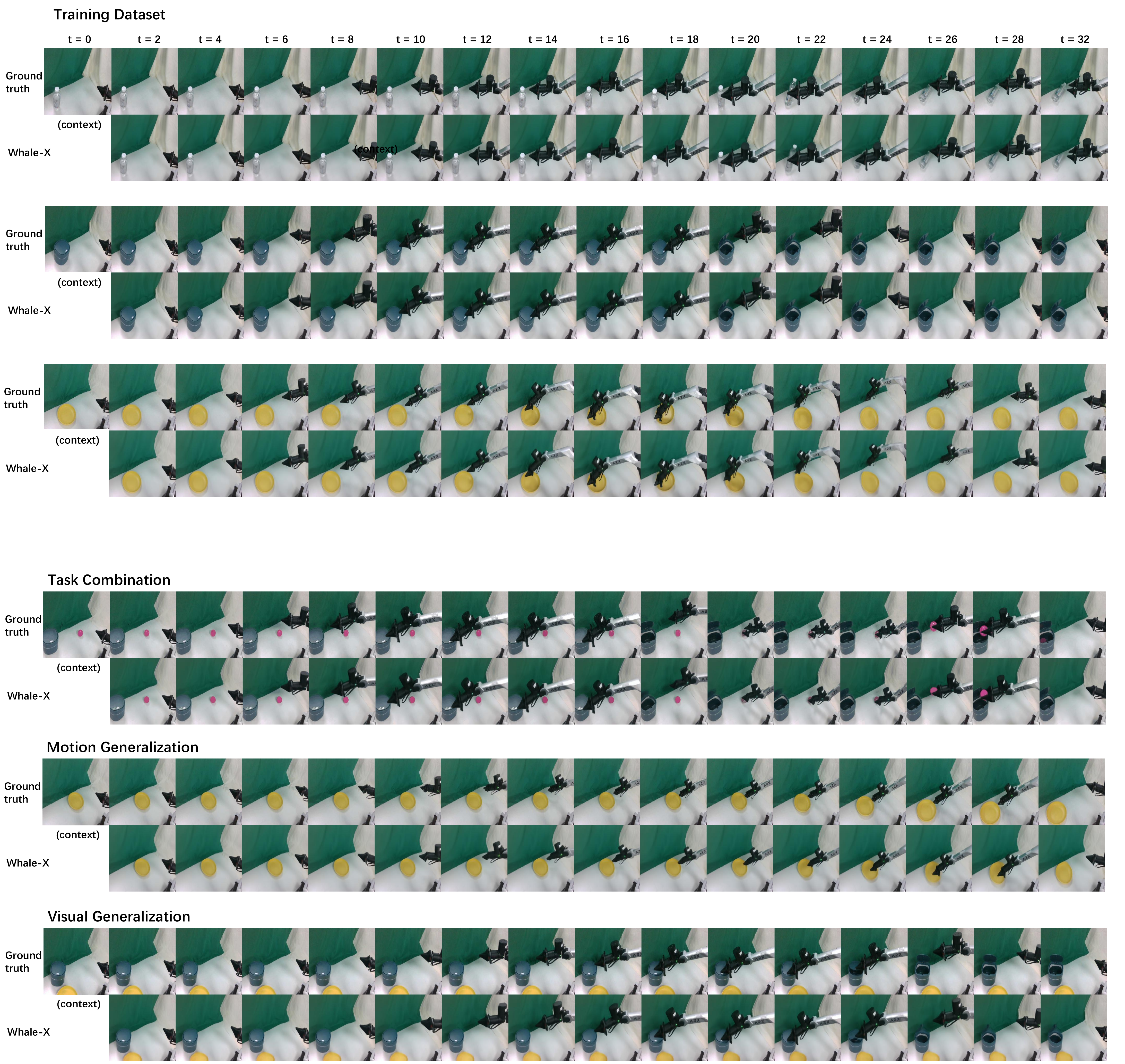}
    \caption{Additional qualitative evaluation on the Real-world tasks. }
    \label{fig:qualitative_realworld}
\end{figure*}

\paragraph{Qualitative Evaluation on Real-world Task}
Figure~\ref{fig:qualitative_realworld} shows the qualitative evaluation results of Whale-X on Real-world Tasks. Whale-X demonstrates strong generalizability in terms of motion, visualization, and task combination.
\label{vis:real_world}

\begin{figure*}[!t]
    \centering
    \vspace{-2mm}
    
    \begin{minipage}[b]{0.31\linewidth}
        \centering
        \includegraphics[width=\linewidth]{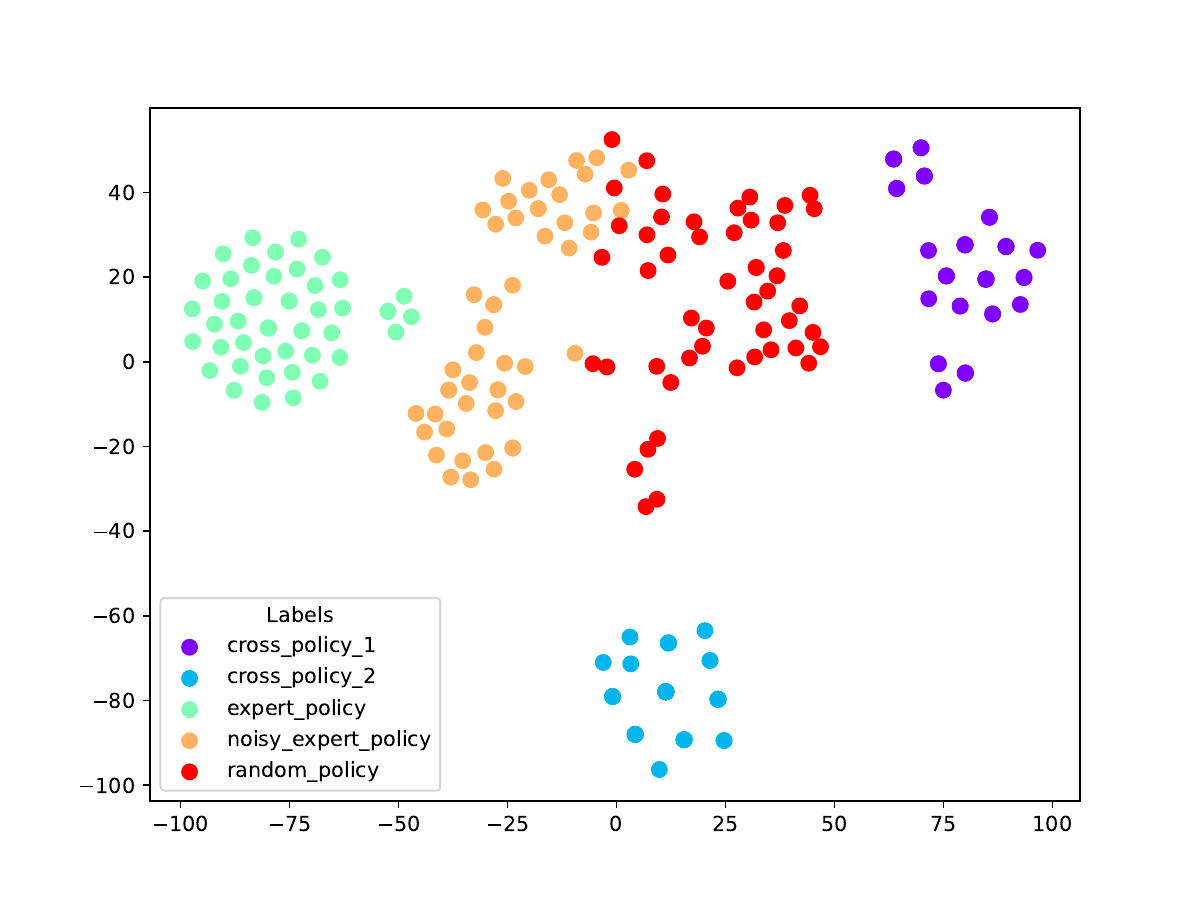}
        \subcaption{Same task different policies}
        \label{fig:same_task_diff_policies}
    \end{minipage}
    \begin{minipage}[b]{0.31\linewidth}
        \centering
        \includegraphics[width=\linewidth]{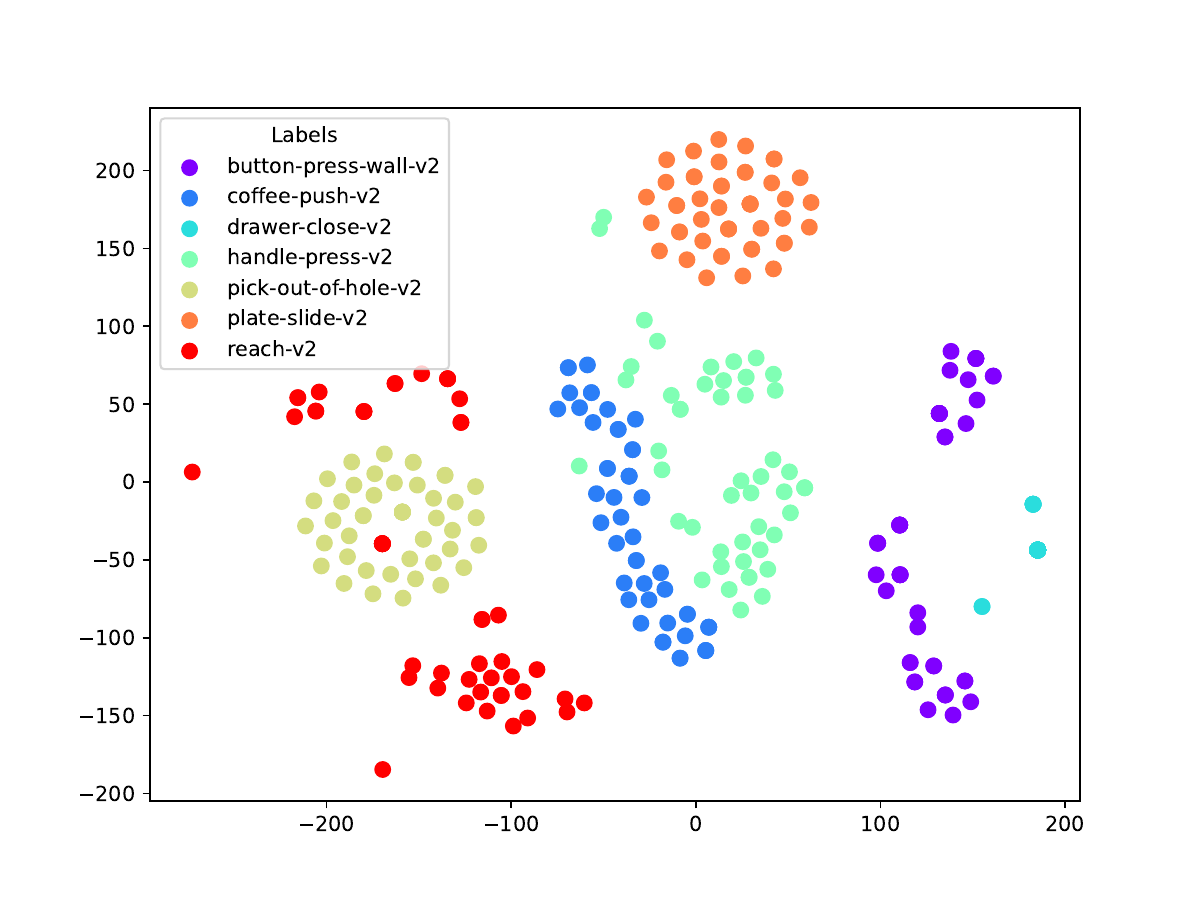}
        \subcaption{Different tasks expert policies}
        \label{fig:diff_tasks_expert_policies}
    \end{minipage}
    \begin{minipage}[b]{0.31\linewidth}
        \centering
        \includegraphics[width=\linewidth]{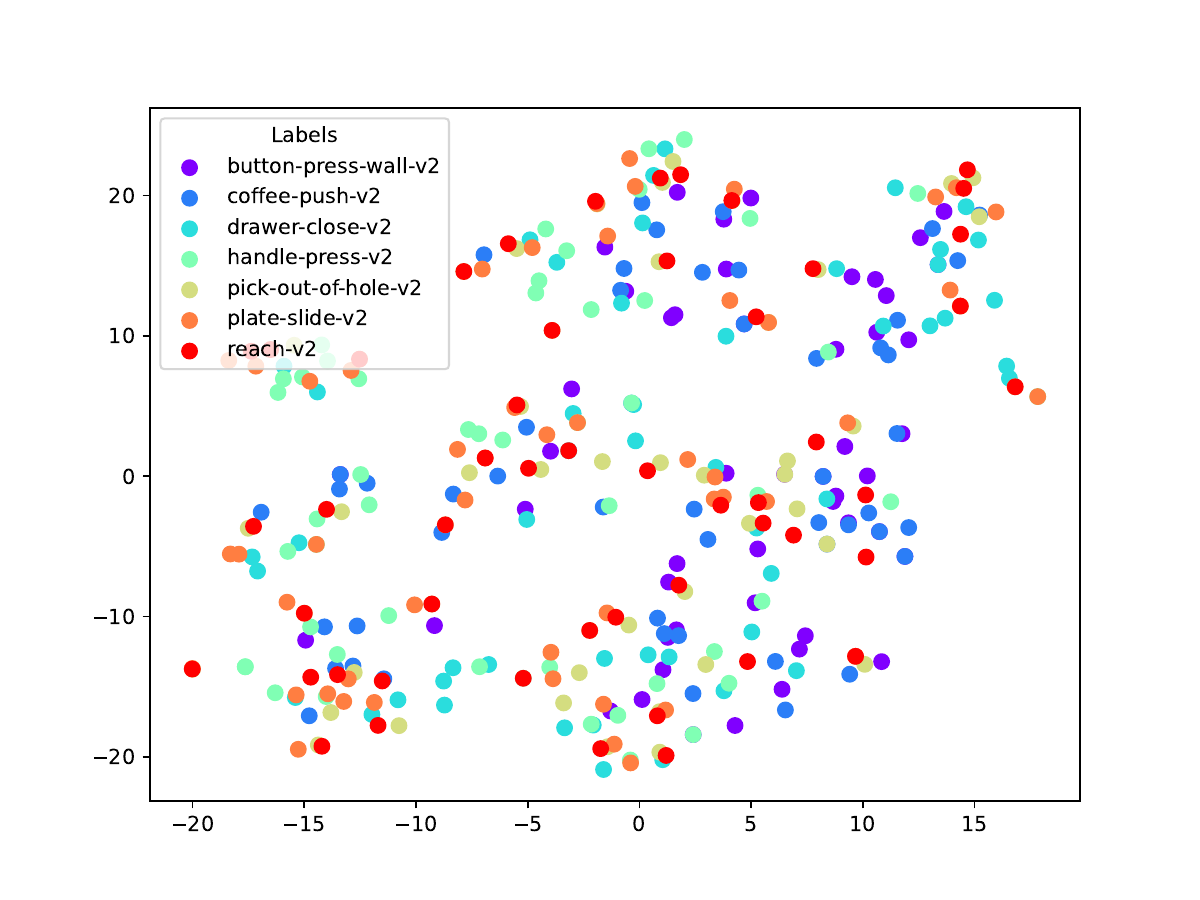}
        \subcaption{Different tasks random policies}
        \label{fig:diff_tasks_rnd_policies}
    \end{minipage}
    
    \caption{The behavior embedding visualization via t-SNE~\citep{tsne}. The different colors denote different policies in the same task (\ref{fig:same_task_diff_policies}) and expert policies in different tasks (\ref{fig:diff_tasks_expert_policies}) or random policies in different tasks (\ref{fig:diff_tasks_rnd_policies}).}
    \vspace{-3mm}
    \label{fig:embedding visualization}
\end{figure*}

\subsection{Uncertainty Estimation Visualization} 
\label{app:uncertainty_quanti}
\begin{figure*}[htbp]
    \centering
    \includegraphics[width=1\linewidth]{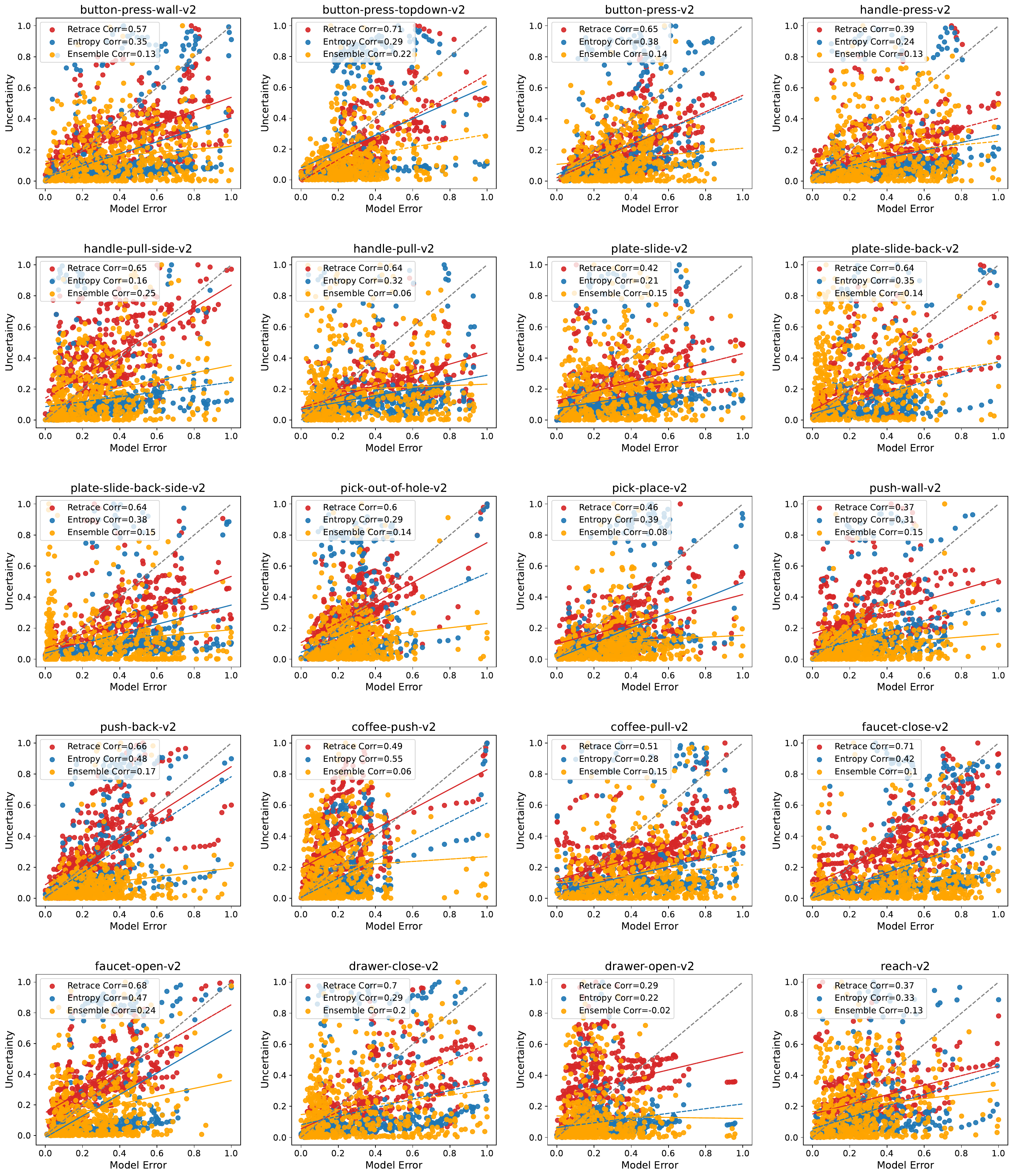}
    \caption{The correlation between the estimated uncertainty and the perceptual model error.}
    \label{fig:uncertainty_qunatification}
\end{figure*}

Figure \ref{fig:uncertainty_qunatification} presents model error prediction results across all 20 tasks, illustrating that the retracing-rollout method consistently outperforms both entropy-based and ensemble-based methods in accuracy.

\subsection{Behavior Embedding Visualization}
\label{app:embed_visual}
Figure \ref{fig:embedding visualization} shows the visualization results of behavior embedding analysis.

\section{GPT-4o Evaluation Details}
\label{app:GPT-4o evaluation}
\subsection{Q\&A Example}
\begin{figure*}[!t]
    \centering
    \vspace{-5mm}
    \includegraphics[width=1\linewidth]{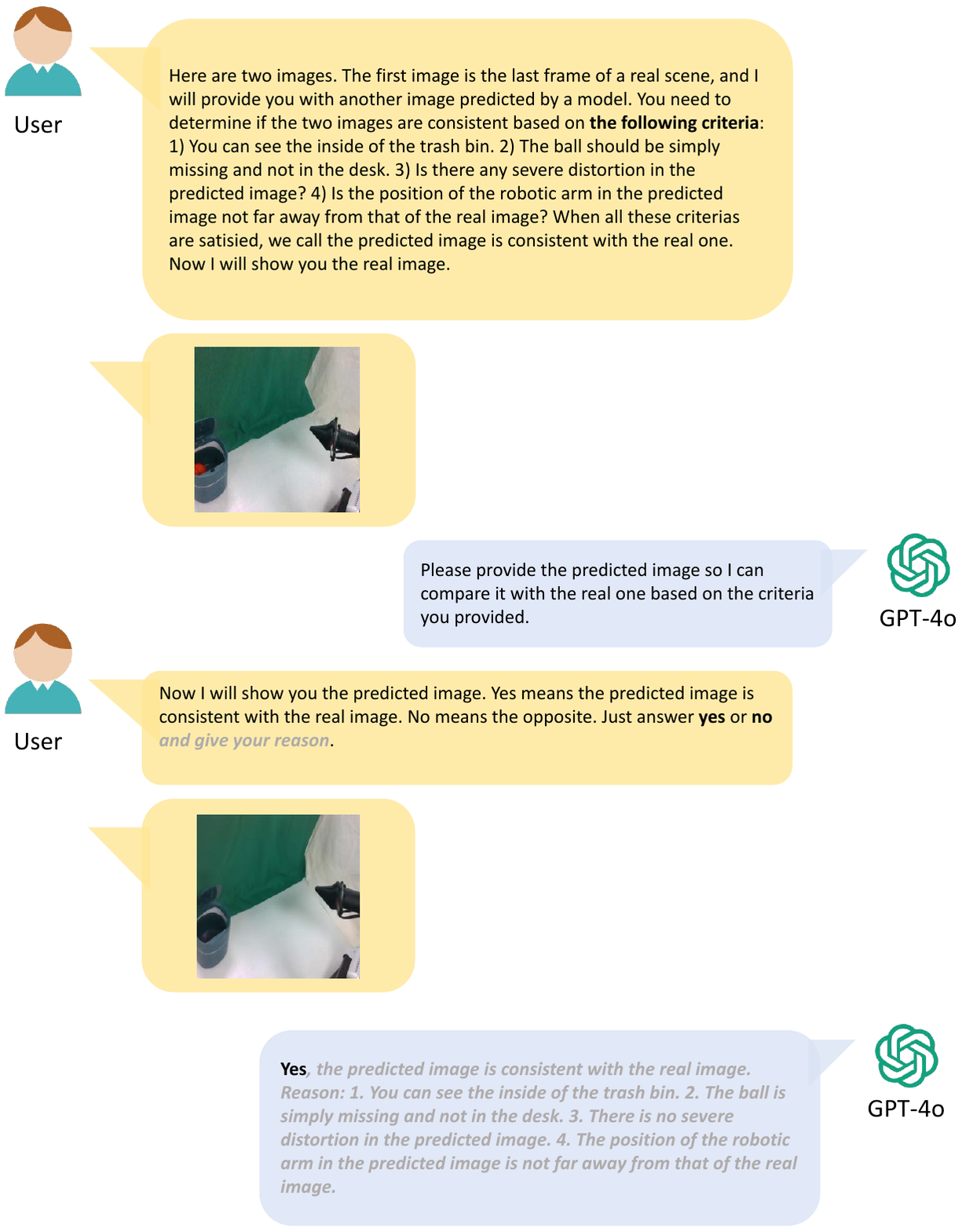}
    \vspace{-13mm}
    \caption{The illustration of a Q\&A example using GPT-4o for evaluating the world model's consistency rate.}
    \label{fig:fig_plots_diaglogue}
\end{figure*}

We use the large vision language model GPT-4o for evaluation in the physical robot experiment. Generally, we input the real final frame and the model-generated final frame to GPT-4o, using natural language dialogue to enable GPT-4o to assess whether the generated errors in key information such as the robot arm's position and the status of interactive objects fall within an acceptable range, thus determining whether the generated results are consistent with reality. Figure~\ref{fig:fig_plots_diaglogue} shows one of our dialogue examples with GPT-4o. We use multi-turn dialogue to enable the model to easily process and infer information from images.

\subsection{All GPT-4o Prompts}
Table~\ref{tab:all_prompts_x} contains all the prompts we used with GPT-4o for evaluation in unseen scenarios. The prompts evaluate various criteria by listing factors such as the robotic arm's position and the status of interactive objects.

\begin{table}[htbp]
\centering
\begin{tabular}{p{3cm}|p{12cm}}
\toprule
\multicolumn{1}{c|}{\textbf{Task}} & \multicolumn{1}{c}{\textbf{Prompt}}\\
\midrule
Visual Generalization & Here are two images. The first image is the last frame of a real scene, and I will provide you with another image predicted by a model. The task is to open the trash bin under an unseen background. The trash bin is on the left side of the desk and is closed at the beginning. You need to determine if the two images are consistent based on the following criteria: 1) You can see the inside of the trash bin.  2) Is the predicted image clear? When all these criteria are satisfied, we call the predicted image consistent with the real one. Now I will show you the real image. \\
\midrule
Motion Generalization & Here are two images. The first image is the last frame of a real scene, and I will provide you with another image predicted by a model. You need to determine if the two images are consistent based on the following criteria: 1) Is the plate's position on the left side of the image? 2) Does the plate disappear in the predicted image? 3) Is the predicted image clear? 4) Is the robot arm still present in the predicted image? 5) Does the position of the robot arm in the predicted image match that of the real image? When all these criteria are satisfied, we call the predicted image consistent with the real one. Now I will show you the real image. \\
\midrule
Task Generalization & Here are two images. The first image is the last frame of a real scene, and I will provide you with another image predicted by a model. You need to determine if the two images are consistent based on the following criteria: 1) You can see the inside of the trash bin. 2) The ball should be simply missing and not on the desk. 3) Is there any severe distortion in the predicted image? 4) Is the position of the robot arm in the predicted image not far away from that of the real image? When all these criteria are satisfied, we call the predicted image consistent with the real one. Now I will show you the real image.\\
\bottomrule
\end{tabular}
\caption{The prompt used for 3 unseen tasks.}
\label{tab:all_prompts_x}
\end{table}

\subsection{More Evaluation Results}
\begin{figure*}[h!]
    \centering
    \includegraphics[width=0.95\linewidth]{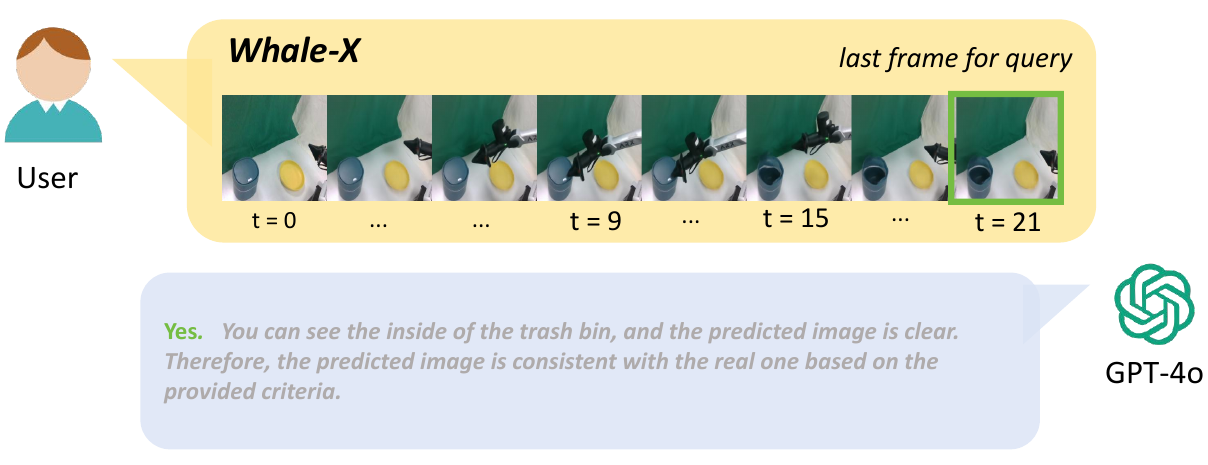}
    \caption{The example of GPT-4o evaluation for Whale-X on the Visual Generalization Task.}
    \label{fig:fig_plots_example001_001}
    \vspace{-5mm}
\end{figure*}

\begin{figure*}[h!]
    \centering
    \includegraphics[width=0.95\linewidth]{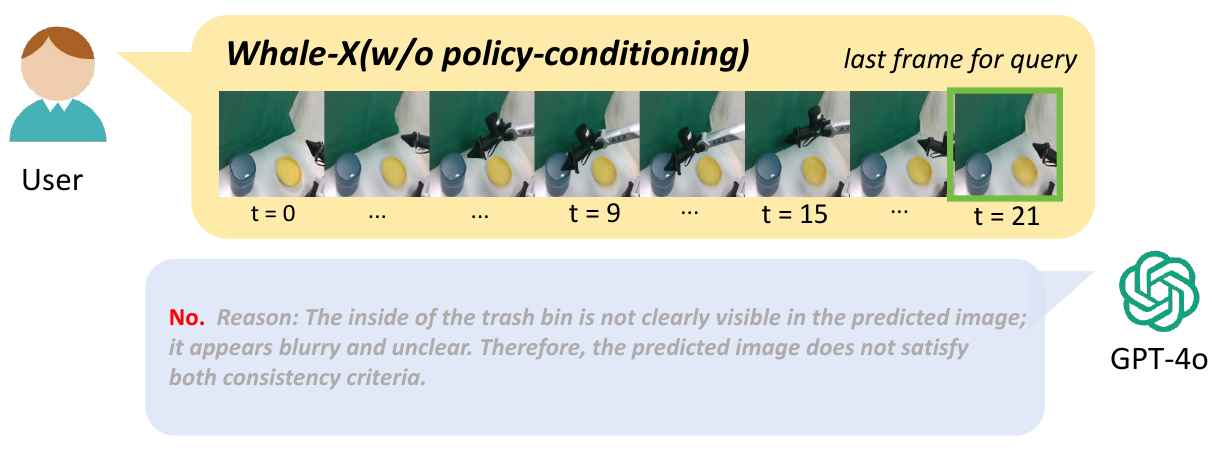}
    \caption{The example of GPT-4o evaluation for Whale-X(w/o policy-conditioning) on the Visual Generalization Task.}
    \label{fig:fig_plots_example001_002}
    \vspace{-5mm}
\end{figure*}

\begin{figure*}[h!]
    \centering
    \includegraphics[width=0.95\linewidth]{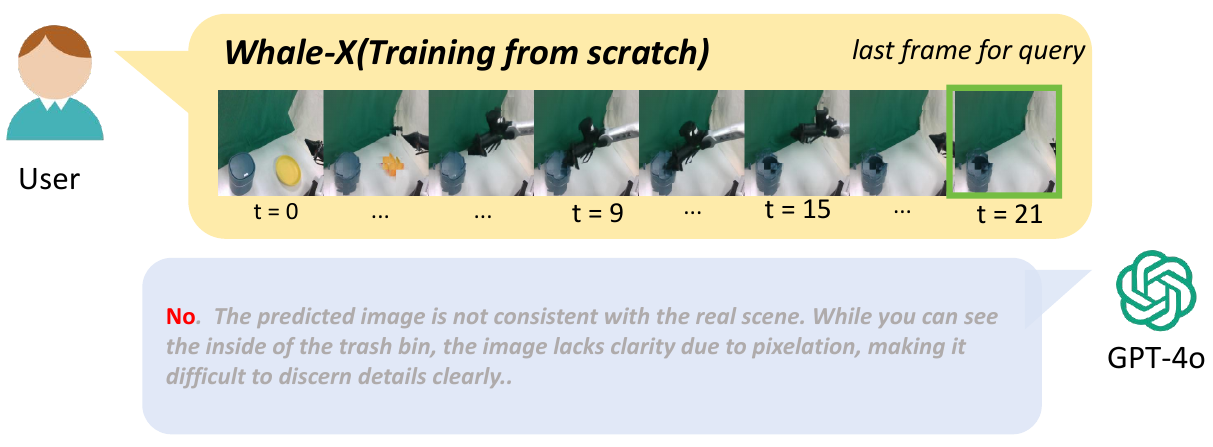}
    \caption{The example of GPT-4o evaluation for Whale-X(Training from scratch) on the Visual Generalization Task.}
    \label{fig:fig_plots_example002_002}
    \vspace{-5mm}
\end{figure*}

\begin{figure*}[h!]
    \centering
    \includegraphics[width=0.95\linewidth]{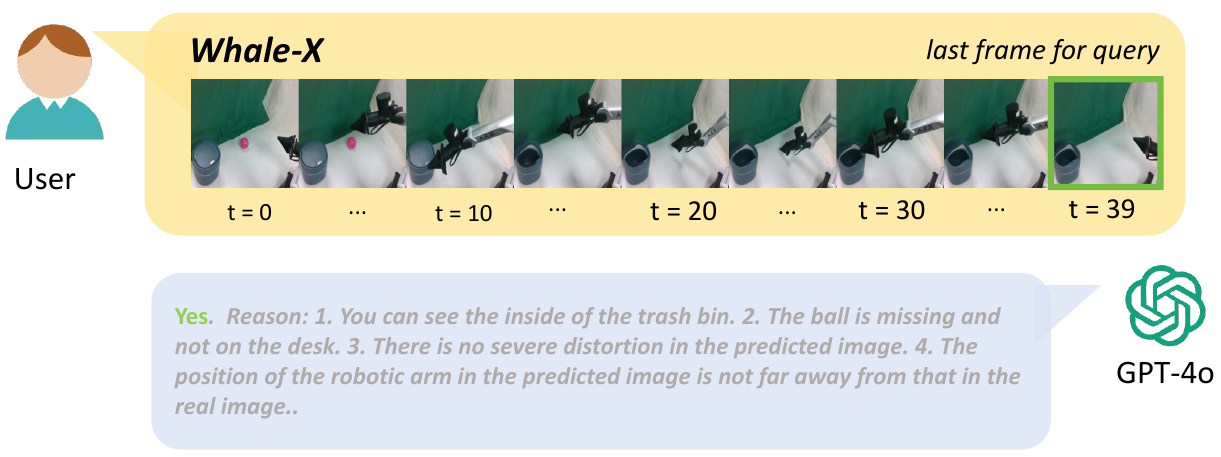}
    \caption{The example of GPT-4o evaluation for Whale-X on the Task Generalization Task.}
    \label{fig:fig_plots_example005_001}
\end{figure*}

\begin{figure*}[h!]
    \centering
    \includegraphics[width=0.95\linewidth]{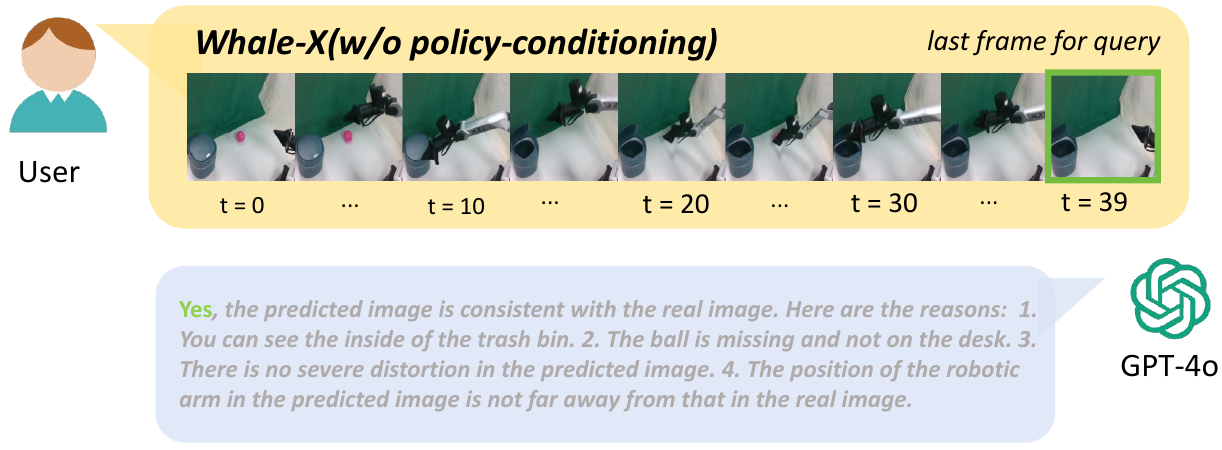}
    \caption{The example of GPT-4o evaluation for Whale-X(w/o policy-conditioning) on the Task Generalization Task.}
    \label{fig:fig_plots_example005_002}
\end{figure*}

\begin{figure*}[h!]
    \centering
    \includegraphics[width=0.95\linewidth]{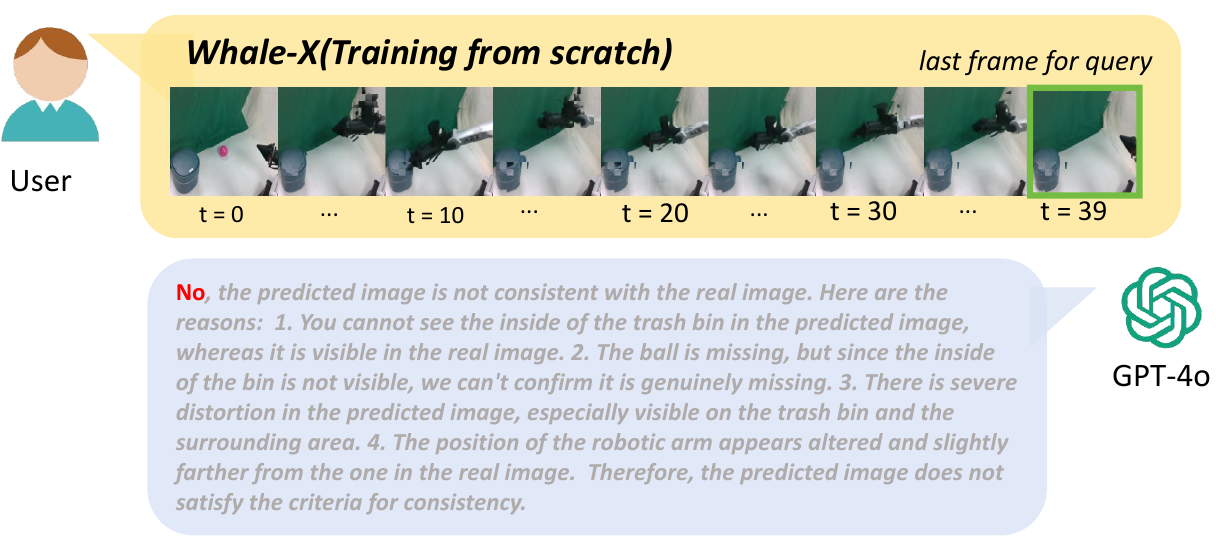}
    \caption{The example of GPT-4o evaluation for Whale-X(Training from scratch) on the Task Generalization Task.}
    \label{fig:fig_plots_example006_002}
\end{figure*}

\begin{figure*}[h!]
    \centering
    \includegraphics[width=0.95\linewidth]{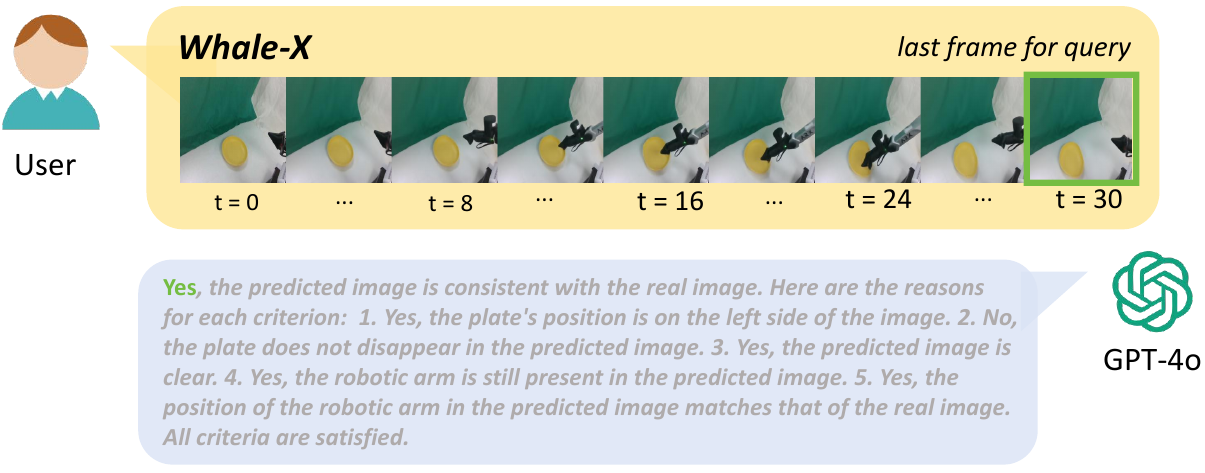}
    \caption{The example of GPT-4o evaluation for Whale-X on the Motion Generalization Task.}
    \label{fig:fig_plots_example003_001}
\end{figure*}

\begin{figure*}[h!]
    \centering
    \includegraphics[width=0.95\linewidth]{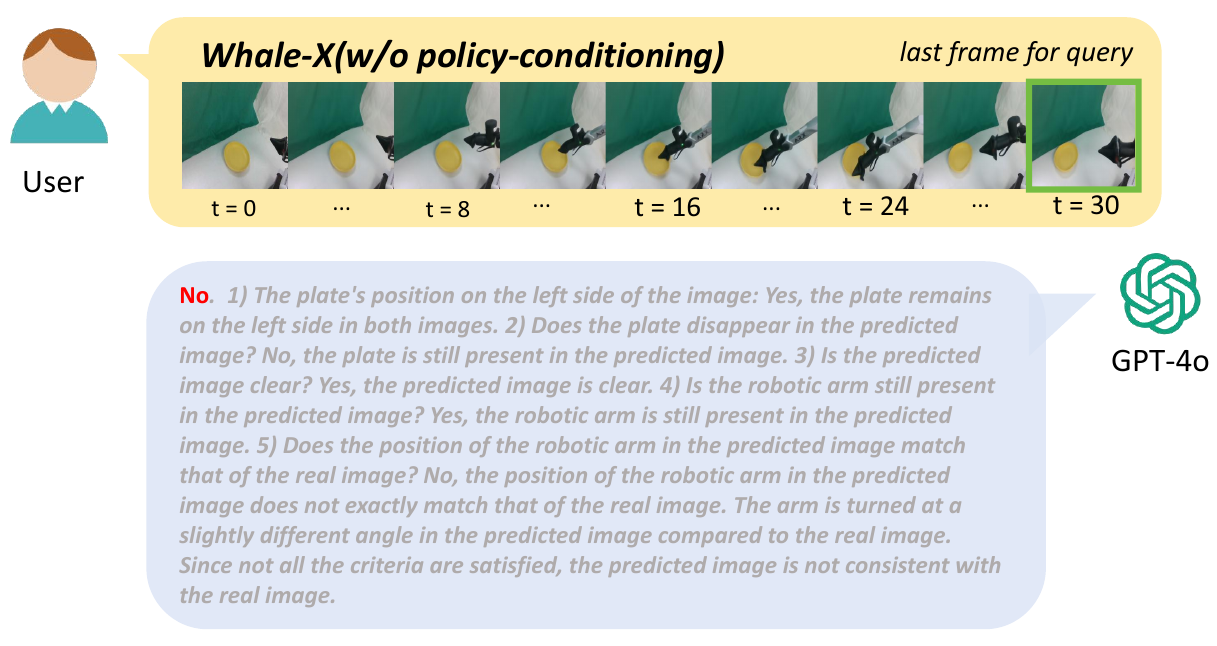}
    \caption{The example of GPT-4o evaluation for Whale-X(w/o policy-conditioning) on the Motion Generalization Task.}
    \label{fig:fig_plots_example003_002}
\end{figure*}

\begin{figure*}[h!]
    \centering
    \includegraphics[width=0.95\linewidth]{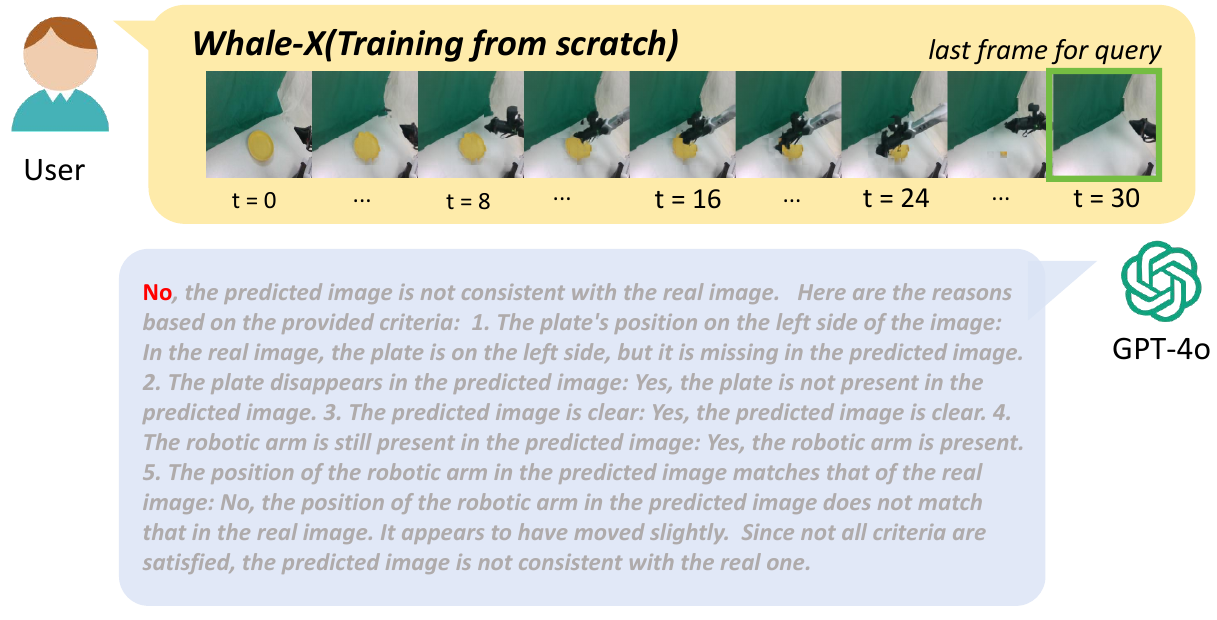}
    \caption{The example of GPT-4o evaluation for Whale-X(Training from scratch) on the Motion Generalization Task.}
    \label{fig:fig_plots_example004_002}
\end{figure*}

Figure~\ref{fig:fig_plots_example001_001}~\ref{fig:fig_plots_example001_002}~\ref{fig:fig_plots_example002_002} show the evaluation results for Whale-X, Whale-X (w/o behavior-conditioning), and Whale-X (training from scratch) on the Visual Generalization task. Figure~\ref{fig:fig_plots_example003_001}~\ref{fig:fig_plots_example003_002}~\ref{fig:fig_plots_example004_002} show the evaluation results for Whale-X, Whale-X (w/o behavior-conditioning), and Whale-X (training from scratch) on the Motion Generalization task. Figure~\ref{fig:fig_plots_example005_001}~\ref{fig:fig_plots_example005_002}~\ref{fig:fig_plots_example006_002} show the evaluation results for Whale-X, Whale-X (w/o behavior-conditioning), and Whale-X (training from scratch) on the Task Generalization task.

\section{Computational Resources}

Our models for simulated Meta-World tasks are trained and evaluated on a single RTX 4090 GPU platform, while Whale-X for real-world robot manipulation is trained and evaluated using 8 RTX 4090 GPUs. On simulated tasks, it is approximately 2 days for tokenizer training, 8 hours for behavior-conditioning model training, and 1 day for dynamics model training, totaling around 3 days. Pre-training a Whale-X-base model using 8 RTX-4090 GPUs takes about 10 days in total. Specifically, tokenizer training requires 6 days, dynamics model training takes 3 days, and behavior-conditioning model training takes less than 1 day.
\end{document}